\documentclass[journal]{IEEEtran}
\usepackage{cite}

\usepackage{color}

\usepackage{amsmath}
\usepackage{algorithm}
\usepackage{algorithmic}
\usepackage{bm}
\usepackage{latexsym}
\usepackage{amsthm}
\usepackage{url}
\usepackage{amsfonts}
\usepackage{amssymb}
\usepackage{color}
\usepackage{indentfirst}
\usepackage{float}
\usepackage{array}
\usepackage{multirow}
\usepackage{graphicx}  
\usepackage{subfig}
\usepackage{booktabs}
\usepackage{multirow}
\usepackage{rotating}
\usepackage{bbding}
\usepackage{chngpage}
\usepackage{makecell}

\makeatletter \@fpsep = 1pt \makeatother
\ifCLASSINFOpdf
\else
\fi
\begin{document}

\title{A Survey on Incomplete Multi-view Clustering}
\author{Jie Wen, Zheng Zhang, \emph{Senior Member}, \emph{IEEE}, Lunke Fei, Bob Zhang, \emph{Senior Member}, \emph{IEEE},\\
Yong Xu, \emph{Senior Member}, \emph{IEEE}, Zhao Zhang, \emph{Senior Member}, \emph{IEEE}, Jinxing Li
\thanks{This work is supported by the Shenzhen Science and Technology Program under Grant no. RCBS20210609103709020 and Shenzhen Fundamental Research Fund under Grant no. JCYJ20190806142416685. (\emph{Jie Wen and Zheng Zhang contribute equally.})}
\thanks{Jie Wen, Zheng Zhang, Yong Xu, and Jinxing Li are with the Shenzhen Key Laboratory of Visual Object Detection and Recognition, Harbin Institute of Technology, Shenzhen, Shenzhen 518055, China. (Email: jiewen\_pr@126.com; darrenzz219@gmail.com; laterfall@hit.edu.cn; lijinxing@cuhk.edu.cn)}
\thanks{Lunke Fei is with the School of Computer Science and Technology, Guangdong University of Technology, Guangzhou, China. (Email: flksxm@126.com)}
\thanks{Bob Zhang is with the Department of Computer and Information Science, University of Macau, Taipa, Macau. (Email: bobzhang@um.edu.mo).}
\thanks{Zhao Zhang is with the School of Computer Science \& School of Artificial Intelligence, Hefei University of Technology, Hefei 230000, China. (Email: cszzhang@gmail)}
\thanks{Corresponding authors: Bob Zhang (Email: bobzhang@um.edu.mo) and Yong Xu (Email: laterfall@hit.edu.cn).}
}

\markboth{IEEE Transactions on Systems, Man, and Cybernetics: Systems}%
{Shell \MakeLowercase{\textit{et al.}}: Bare Demo of IEEEtran.cls for IEEE Journals}
\maketitle

\begin{abstract}
Conventional multi-view clustering seeks to partition data into respective groups based on the assumption that all views are fully observed. However, in practical applications, such as disease diagnosis, multimedia analysis, and recommendation system, it is common to observe that not all views of samples are available in many cases, which leads to the failure of the conventional multi-view clustering methods. Clustering on such incomplete multi-view data is referred to as incomplete multi-view clustering. In view of the promising application prospects, the research of incomplete multi-view clustering has noticeable advances in recent years. However, there is no survey to summarize the current progresses and point out the future research directions. To this end, we review the recent studies of incomplete multi-view clustering. Importantly, we provide some frameworks to unify the corresponding incomplete multi-view clustering methods, and make an in-depth comparative analysis for some representative methods from theoretical and experimental perspectives. Finally, some open problems in the incomplete multi-view clustering field are offered for researchers. The related codes are released at {\url{https://github.com/DarrenZZhang/Survey_IMC}}.
\end{abstract}

\begin{IEEEkeywords}
Incomplete multi-view clustering, missing views, multi-view learning, data mining.
\end{IEEEkeywords}
\IEEEpeerreviewmaketitle

\section{Introduction}\label{sec:introduction}
\IEEEPARstart{I}{n} recent years, multi-view data collected from diverse sources can be seen everywhere \cite{jack2008alzheimer,yan2019protein,bhadra2017identification,zhou2019dual}. As shown in Fig. 1, doctors usually combine the information from magnetic resonance imaging (MRI), positron emission tomography (PET), and cerebrospinal fluid (CSF), for diagnosing Alzheimer`s disease \cite{jack2008alzheimer}. Features obtained by different feature extractors, such as texture, color, and local binary patterns, can be also regarded as different views of an image \cite{cai2013heterogeneous}. Generally speaking, different views not only contain complementary information and consistent information, but also have many redundant information and inconsistent information \cite{shao2015multiple,guo2016multiview}. This indicates that it is not a good approach to stack different views into a long vector or treat these views individually in practical applications. To address this issue, the research on multi-view learning appears, which aims to jointly explore the information of multiple views for different tasks \cite{pan2021multi,lin2021multi}.

\begin{figure*}[ht]
\centering
\includegraphics[width=4.5in,height=0.8in]{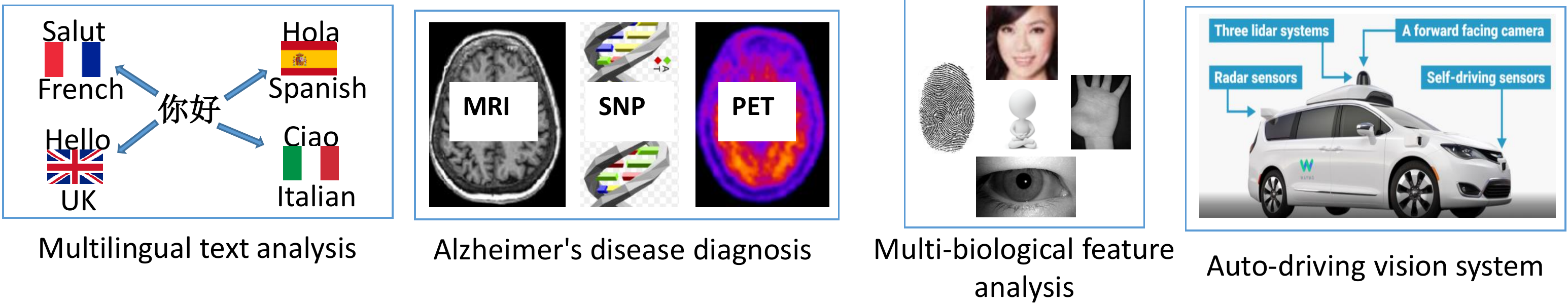}
\caption{Some examples of multi-view data \cite{chao2017survey}.}
\label{fig:fig1}
\end{figure*}

As an unsupervised multi-view learning method, multi-view clustering has received a lot of attention in fields of computer vision, bioinformatics, and natural language processing \cite{chao2017survey,liu2017spectral,huang2019multi,kang2020multi}. It focuses on partitioning a set of data samples into different groups according to the underlying information of multiple views. In the past decades, plenty of multi-view clustering methods have been proposed based on a common assumption that all views are fully observed, where some of these methods are summarized in \cite{chao2017survey,yang2018multi,xu2013survey,kang2021structured,li2020bipartite}. However, in many practical applications today, the multi-view data to be processed are usually incomplete, \emph{i.e.}, some of the samples do not have all views. For instance, as stated in \cite{liu2017view}, owing to the patient dropout and poor data quality, some images corresponding to the PET and CSF are not available for some individuals, which generates an incomplete multi-view data of Alzheimer`s disease. When some views are missing, the natural alignment property of multiple views will be seriously broken, which is harmful to the excavation of the complementary information and consistent information. In addition, the absence of views leads to a serious of information loss and aggravates the information imbalance among different views. These indicate that the incomplete learning problem is challenging. In fields of multi-view clustering, clustering on such incomplete multi-view data is referred to as \emph{incomplete multi-view clustering (IMC)}.

For IMC, two naive approaches are: 1) removing the samples with missing views and then performing clustering on the remaining samples with fully observed views. 2) Setting the missing views as 0 or average instance and then processing the data by conventional multi-view clustering methods \cite{li2014partial,zhao2016incomplete}. The first approach goes against the purpose of clustering which aims at partitioning all data points into their respective clusters \cite{li2014partial}. For the second approach, the filled missing instances will play a negative role in the clustering process since all missing instances filled in the same vector will be naturally segmented to the same cluster. Besides this, the experimental results in Section VII also show that the second approach achieves very bad performances, especially for the case with a high missing-view rate. In the past years, many advanced methods, such as cascaded residual autoencoder (CRA) \cite{tran2017missing} and missing view imputation with generative adversarial networks (VIGAN) \cite{shang2017vigan}, are proposed for missing view recovery. CRA stacks many residual autoencoders and recovers the missing views by minimizing the residual between the prediction and original data. VIGAN combines the denoising autoencoder and generative adversarial networks (GAN), where denoising autoencoder is used to reconstruct the missing views according to the outputs of GAN. A limitation of VIGAN is that this method is not suitable to process the incomplete data with more than three views. Moreover, recovering the missing views is a promising approach to address the incomplete learning problem but the performance will be highly dependent on the quality of the recovered data.

\textbf{Motivation and objectives}: In recent years, IMC has received more and more attention and many effective methods have been proposed to address such a challenging problem \cite{wen2018incompleteECCV,wang2020generative,wen2018incompleteTCYB,wen2019unified,zhou2019consensus,liu2019multiple,liu2019efficient,wen2020Generalized,guo2019anchors,zhu2018localized}. However, there is no survey to systematically summarize the current progresses of IMC. The main purpose of this paper is to provide a comprehensive study on IMC and to provide a good start to newcomers who are interested in IMC and its related areas. In this paper, almost all of the existing state-of-the-art IMC methods are summarized. Besides this, some representative IMC methods are selected and fairly compared, which lets the readers intuitively observe the clustering performance of these methods. Moreover, some open problems that still have not been solved well are offered for researchers.

\textbf{Categorization of the existing methods}: The existing IMC methods can be categorized from different perspectives. For example, from the viewpoint of `missing view recovery', the existing IMC methods can be categorized into two groups, where the one focuses on addressing the incomplete learning problem by recovering the missing views or missing connections among samples, and the other one does not recover the missing information \emph{w.r.t.} the missing views but just focuses on the partially aligned information among the available views. From the viewpoint of the main methodologies and learning models exploited in these methods, we can divide the existing IMC methods into four categories, \emph{i.e.}, matrix factorization (MF) based IMC, kernel learning based IMC, graph learning based IMC, and deep learning based IMC. To summarize more IMC methods and provide an intuitive comparison, we will analyze the existing IMC methods according to the second categorization scheme in this paper. Specifically, in our work, we regard all IMC methods that exploit the deep network as deep learning based IMC since these methods are very rare in comparison with the other methods. MF based IMC generally seeks to decompose the multi-view data into the consensus representation shared by all views. Kernel learning based IMC tries to obtain the consensus representation from the incomplete kernel data. Graph learning based IMC is based on the spectral clustering, which aims to obtain a consensus graph (or several view-specific graphs) from incomplete multi-view data or calculate the consensus representation from the incomplete graphs directly.

\textbf{Organization of the paper}: The structure of this paper is shown in Fig.2. We first introduce some preliminary concepts in section II. In sections III-VI, we summarize and analyze the four kinds of IMC methods, respectively. In section VII, we conduct several experiments to compare and analyze the representative IMC methods. In section VIII, we offer the conclusions and some open problems. More experiments are presented in the supplementary material.
\begin{figure}[ht]
\centering
\includegraphics[width=3.5in,height=2.2in]{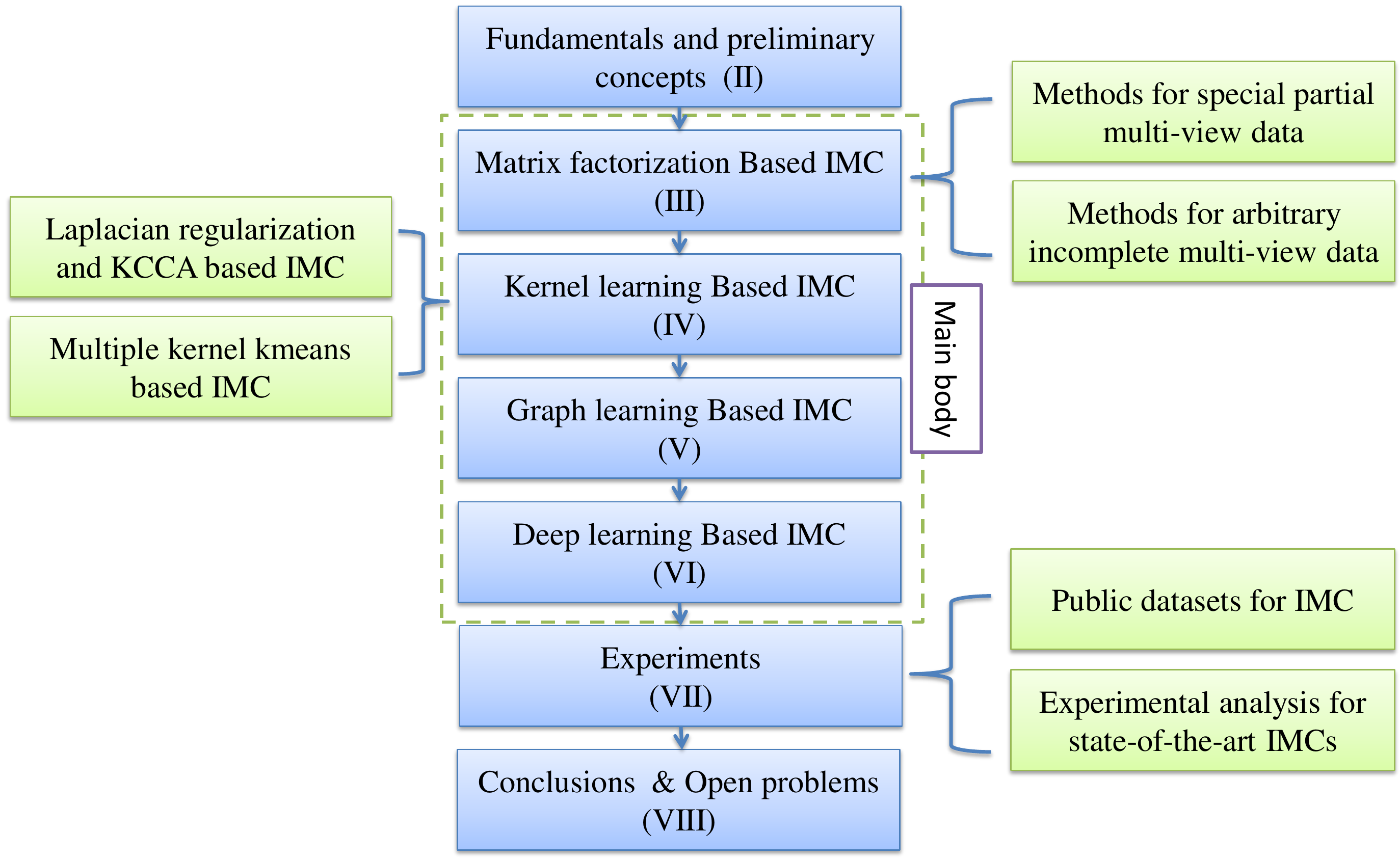}
\caption{The structure of the paper.}
\label{fig:fig2}
\end{figure}
\section{Fundamentals and preliminary concepts}
\subsection{Notations}
Owing to the space limitation, we move the detailed notations used through the paper to Table V of the supplementary file. For a matrix $A \in R^{m\times n}$, its $l_1$ norm, $l_F$ norm, and $l_{2,1}$ norm are defined as ${\left\| A \right\|_1} = \sum\limits_{i = 1}^m {\sum\limits_{j = 1}^n {\left| {{A_{i,j}}} \right|} } $, $\left\| A \right\|_F^2 = \sum\limits_{j = 1}^n {\sum\limits_{i = 1}^m {A_{i,j}^2} } $, and ${\left\| A \right\|_{2,1}} = \sum\limits_{i = 1}^m {{{\left( {\sum\limits_{j = 1}^n {A_{i,j}^2} } \right)}^{{1 \mathord{\left/
 {\vphantom {1 2}} \right.
 \kern-\nulldelimiterspace} 2}}}} $, respectively. $A \ge 0$ means that all elements of matrix $A$ are non-negative. $A^T$ is the transpose matrix of $A$. $Tr\left( A \right) = \sum\limits_{i = 1}^n {{A_{i,i}}} $ is the trace of matrix $A \in R^{n \times n}$. $\left| Z \right|$ is the element-wise absolute matrix of $Z$ with all elements as $\left| {{Z_{i,j}}} \right|$.

\subsection{Basic background on single-view/multi-view clustering}
Considering that many IMC methods are derived from the MF based multi-view clustering and multi-view spectral clustering, we briefly introduce some related basic knowledges.

\textbf{MF based multi-view clustering}: For multi-view data ${T^{(v)}\in R^{m_v \times n}}$, a basic assumption is that different views have the same distribution of labels. This is also the so-called semantic consistency of multiple views. A naive MF based approach is to obtain the consensus representation $P$ as follows \cite{liu2013multi}:
\begin{equation}\small
\begin{array}{l}
\mathop {\min }\limits_{\left\{ {{U^{\left( v \right)}},{\bar P^{\left( v \right)}}} \right\}_{v = 1}^l,P} \sum\limits_{v = 1}^l \begin{array}{l}
\left\| {{T^{\left( v \right)}} - {U^{\left( v \right)}}{\bar P^{\left( v \right)}}} \right\|_F^2 + {\beta _v} \left\| {{\bar P^{\left( v \right)}} - P} \right\|_F^2\\
 + \varphi \left( {{U^{\left( v \right)}},{\bar P^{\left( v \right)}},P} \right)
\end{array} \\
s.t.\left\{ {{U^{\left( v \right)}},{\bar P^{\left( v \right)}},P} \right\} \in \phi
\end{array}
\end{equation}
where $\phi$ and $\varphi \left( {{U^{\left( v \right)}},{\bar P^{\left( v \right)}},P} \right)$ are the boundary constraint and regularization constraint, respectively.

\textbf{Multi-view spectral clustering}: Similar to MF based methods, multi-view spectral clustering also aims to obtain the consensus representation of multi-view data. Their difference is that multi-view spectral clustering seeks to realize such a goal by exploring the information of all graphs jointly, where a naive framework can be formulated as follows \cite{kumar2011co}:
\begin{equation}\small
\begin{array}{l}
\mathop {\min }\limits_{\left\{ {{{\bar P}^{\left( v \right)}}} \right\}_{v = 1}^l,P} \sum\limits_{v = 1}^l {Tr\left( {{{\bar P}^{\left( v \right)}}{L_{{Z^{(v)}}}}{{\bar P}^{\left( v \right)T}}} \right) + {\beta _v}\varphi \left( {{{\bar P}^{\left( v \right)}},P} \right)} \\
s.t.\left\{ {{{\bar P}^{\left( v \right)}},P} \right\} \in \phi
\end{array}
\end{equation}
where $\phi$ is generally defined as the orthogonal constraint. Regularization constraint $\varphi \left( {{\bar P^{\left( v \right)}},P} \right)$ can be chosen as the co-regularizer $Tr(\bar P^{(v)T}PP^T \bar P^{(v)})$, which pushes all individual representations towards a consensus representation $P$ by minimizing their disagreements.
\subsection{Categorization of the incomplete multi-view data}
In this paper, we divide the incomplete multi-view data into three cases as shown in Figs. 3-5, where Fig.3 and Fig.4 show the special incomplete multi-view data with two views and more than two views, respectively. For these two kinds of incomplete data, samples contain only one view and all views are regarded as the single-view-samples and paired-samples, respectively. Fig.5 shows the incomplete multi-view data with arbitrary missing views.
\begin{figure}[ht!]
\centering
\includegraphics[width=2.5in,height=1in]{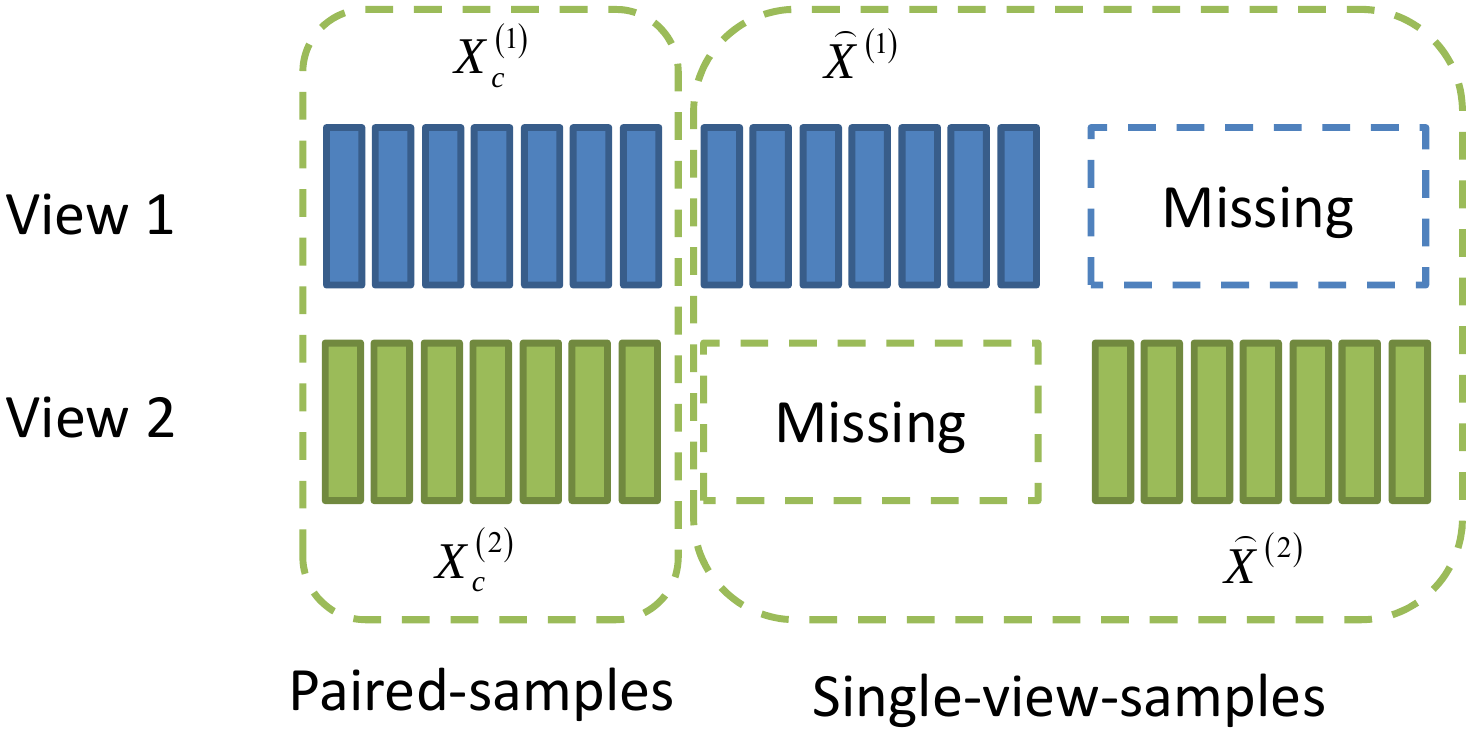}
\caption{Incomplete case of data with two views.}
\label{fig:fig2}
\end{figure}
\begin{figure}[ht!]
\centering
\includegraphics[width=3in,height=1.5in]{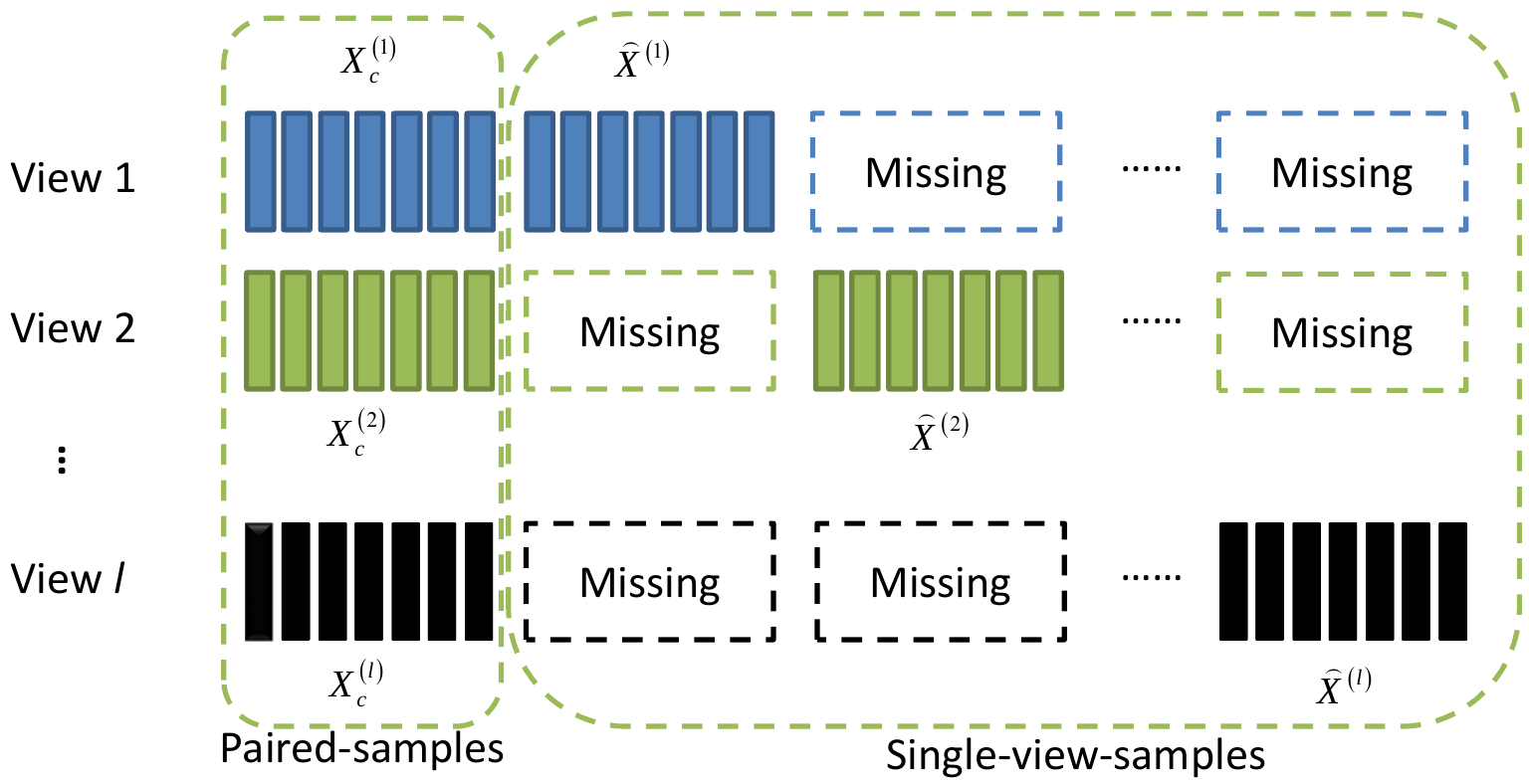}
\caption{Special case of incomplete data with more than two views, which only contains paired-samples and single-view-samples.}
\label{fig:fig3}
\end{figure}
\begin{figure}[ht!]
\centering
\includegraphics[width=2.5in,height=1in]{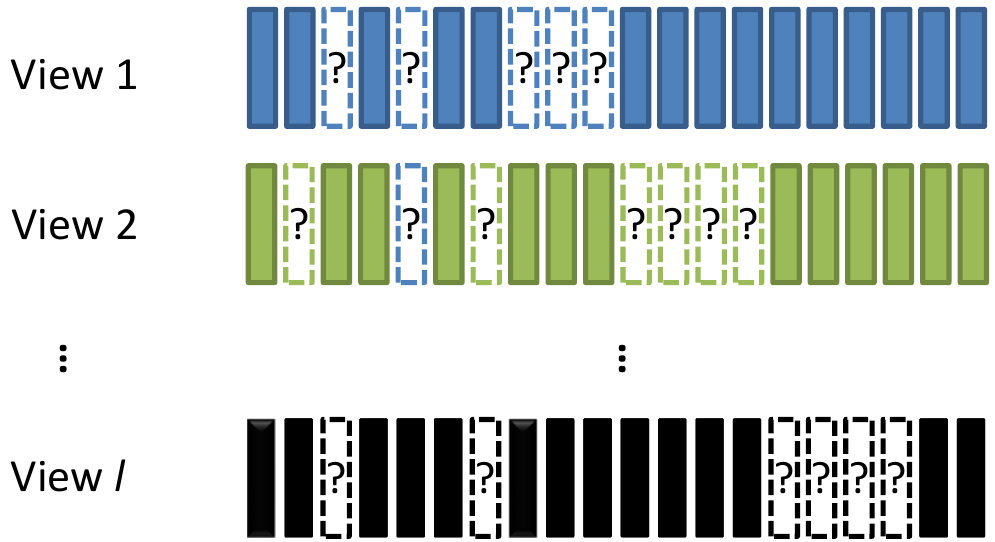}
\caption{Incomplete multi-view data with arbitrary missing instances.}
\label{fig:fig4}
\end{figure}

\section{Matrix factorization based IMC}
From the conventional MF based multi-view clustering framework (1), we can find that it is impossible to obtain the consensus representation directly via such a model when some views are missing. For incomplete multi-view data, the key problem is how to design an IMC model that can obtain a common cluster indicator matrix or consensus representation from the incomplete multi-view data.

In the past years, two approaches are widely considered to design such IMC models based on the MF theory, where one approach aims to explore the consistent information among the partially aligned views, and the other approach focuses on exploring the fully aligned information by recovering the missing views. In this paper, we divide the existing MF based IMC methods into two categories from the viewpoint of application scenario, \emph{i.e.}, MF based methods for special partial multi-view data shown in Figs.3-4 and weighted MF based methods for arbitrary incomplete data shown in Fig.5.
\subsection{MF based methods for special partial multi-view data}
Most of the earliest studies, such as partial multi-view clustering (PMVC) \cite{li2014partial}, incomplete multi-modal grouping (IMG) \cite{zhao2016incomplete}, and partial multi-view subspace clustering (PMSC) \cite{xu2018partial}, take the incomplete multi-view data shown in Fig.3 as an example to design their models. Among these methods, PMVC is a pioneer work, which seeks to obtain the latent common representation $P_c$, $P_s^{(1)}$, and $P_s^{(2)}$, for the paired-samples $\{ X_c^{(v)}\}_{v=1}^2$ and single-view-samples ${{\mathord{\buildrel{\lower3pt\hbox{$\scriptscriptstyle\frown$}} \over X}}^{\left( 1 \right)}}$, ${{\mathord{\buildrel{\lower3pt\hbox{$\scriptscriptstyle\frown$}} \over X} }^{\left( 2 \right)}}$, respectively. However, it ignores the geometric structure of the data, which is important to unsupervised representation learning. IMG \cite{zhao2016incomplete} and PMSC \cite{xu2018partial} are two extensions of PMVC, which further introduce the graph embedding technique to capture the geometric structure. In particular, the above three methods can be unified into the following generalized model, referred to as partial multi-view clustering framework (PMVCF):
\begin{align}\label{eq_5}
&\mathop {\min }\limits_{{P_c},\left\{ {{U^{\left( v \right)}},{P_s^{\left( v \right)}}} \right\}_{v = 1}^l,Z^*} \sum\limits_{v = 1}^l {\left\| {\left[ {X_c^{\left( v \right)},{{\mathord{\buildrel{\lower3pt\hbox{$\scriptscriptstyle\frown$}}
\over X} }^{\left( v \right)}}} \right] - {U^{\left( v \right)}}\left[ {{P_c},{P_s^{\left( v \right)}}} \right]} \right\|_F^2} \nonumber \\
& + \psi \left( {P,Z^*} \right) + \varphi \left( {{U^{\left( v \right)}},{P_s^{\left( v \right)}},{P_c}} \right)\\
&s.t.\left\{ {{P_c},\left\{ {{U^{\left( v \right)}},{P_s^{\left( v \right)}}} \right\}_{v = 1}^l,Z^*} \right\} \in \phi  \nonumber
\end{align}
where $X_c^{\left( v \right)} \in {R^{{m_v} \times {n_c}}}$ denotes the instance set of the paired-samples in the $v$th view. ${\mathord{\buildrel{\lower3pt\hbox{$\scriptscriptstyle\frown$}}
\over X} ^{\left( v \right)}} \in {R^{{m_v} \times {n_{vs}}}}$ denotes the instance set of samples which only exist in the \emph{v}th view, $n_{vs}$ denotes the corresponding instance number. $P = \left[ {{P_c},{P_s^{\left( 1 \right)}}, \ldots ,{P_s^{\left( l \right)}}} \right] \in {R^{c \times n}}$ denotes the consensus representation of all samples, $P_s^{\left( v \right)}$ denotes the new representation of ${\mathord{\buildrel{\lower3pt\hbox{$\scriptscriptstyle\frown$}}
\over X} ^{\left( v \right)}}$. ${U^{\left( v \right)}} \in {R^{{m_v} \times c}}$ is the basis matrix of the $v$th view, $Z^*$ denotes the embedded graph of data. $\psi \left( {P,Z^*} \right)$ and $\varphi \left( {{U^{\left( v \right)}},{P_s^{\left( v \right)}},{P_c}} \right)$ are constraints of the corresponding variables. $\phi$ denotes the boundary constraint of variables.

\textbf{The connections among PMVCF, PMVC, IMG, and PMSC are shown in Fig.6, which mainly include:} 1) PMVC, IMG, and PMSC are all based on a partial multi-view MF model, \emph{i.e.}, $\min \sum\limits_{v = 1}^l {\left\| {\left[ {X_c^{\left( v \right)},{{\mathord{\buildrel{\lower3pt\hbox{$\scriptscriptstyle\frown$}}
\over X} }^{\left( v \right)}}} \right] - {U^{\left( v \right)}}\left[ {{P_c},P_s^{\left( v \right)}} \right]} \right\|_F^2} $. Their motivations on addressing the incomplete learning problem are the same, \emph{i.e.}, using the partially aligned information of paired-samples as constraint to obtain the consensus representation shared by all views. Their main differences are the definitions of constraints $\psi \left( {P,Z^*} \right)$, $\varphi \left( {{U^{\left( v \right)}},{P_s^{\left( v \right)}},{P_c}} \right)$, and boundary constraint $\phi$. 2) Compared with PMVC, IMG and PMSC further explore different kinds of structure information of data by introducing different constraints $\psi \left( {P,Z^*} \right)$.
\begin{figure}[ht]
\centering
\includegraphics[width=3.5in,height=1.6in]{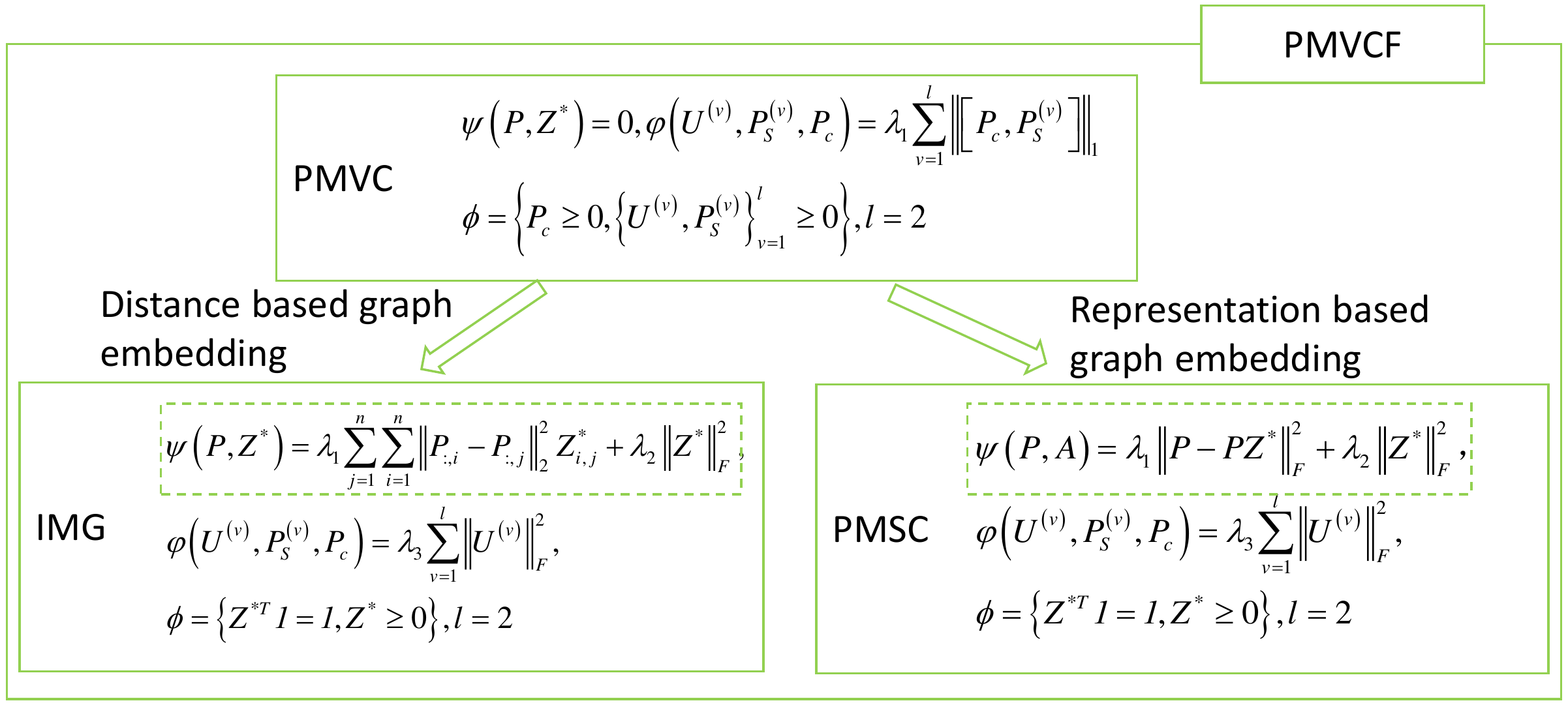}
\caption{Connections among PMVCF, PMVC, IMG, and PMSC.}
\label{fig:fig5}
\end{figure}

Generally, IMG and PMSC have the potential to perform better than PMVC by introducing the manifold constraint. However, their computation complexities are much higher than PMVC. In \cite{wen2018incompleteECCV}, Wen et al. proposed a graph regularized MF based method, referred to as IMC\_GRMF, which integrates the graph regularization and MF into one item as follows:
\begin{equation}\label{eq_4}
\begin{array}{l}
\mathop {\min }\limits_\Upsilon  \sum\limits_{v = 1}^l {\sum\limits_{j = 1}^{{n_c} + {n_v}} {\sum\limits_{i = 1}^{{n_c} + {n_v}} {\left\| {x_i^{\left( v \right)} - {U^{\left( v \right)}}P_{:,j}^{\left( v \right)}} \right\|_2^2Z_{i,j}^{\left( v \right)}} } } \\
 + \sum\limits_{v = 1}^l {\left( {{\lambda _1}\left\| {{P^{\left( v \right)}}{H^{\left( v \right)}} - {P_c}} \right\|_F^2 + {\lambda _2}{{\left\| {{P^{\left( v \right)}}} \right\|}_1}} \right)} \\
s.t.{\kern 1pt} {\kern 1pt} {\kern 1pt} {\kern 1pt} {U^{\left( v \right)T}}{U^{\left( v \right)}} = I
\end{array}
\end{equation}
where $\Upsilon  = \left\{ {{P_c},\left\{ {{P^{\left( v \right)}},{U^{\left( v \right)}}} \right\}_{v = 1}^l} \right\}$. ${Z^{\left( v \right)}} \in {R^{{n_v} \times {n_v}}}$ denotes the k-nearest-neighbor graph pre-constructed from the available instances of the \emph{v}th view. ${H^{\left( v \right)}} \in {R^{{n_v} \times {n_c}}}$ is a pre-constructed binary matrix according to the view-paired-information, where $H_{i,i}^{\left( v \right)} = 1$ ($i = 1, \ldots ,{n_c}$) when the first ${n_c}$ samples are paired samples, otherwise $H_{i,j}^{\left( v \right)} = 0$.

Compared with PMVC, IMG, and PMSC, IMC\_GRMF provides a more general model to handle the data with more than two views as shown in Fig.4. Besides this, IMC\_GRMF exploits an indirect approach to obtain the consensus representation from the latent representations of all views, which provides more freedom for representation learning. However, IMC\_GRMF is inflexible since it requires the feature dimension $c \le \min \left( {{m_1}, \ldots ,{m_l}} \right)$ owing to the orthogonal constraint of the basis matrix.
\subsection{MF based methods for incomplete data with arbitrary missing views}
To handle the incomplete data with arbitrary missing views shown in Fig.5, many weighted MF based incomplete multi-view clustering methods have been proposed, which reduce the negative influence of missing views by imposing some weight matrices pre-constructed from the view-missing information of all views on the MF item. The representative works are multiple incomplete views clustering (MIC) \cite{shao2015multiple}, online multi-view clustering (OMVC) \cite{shao2016online}, doubly aligned incomplete multi-view clustering (DAIMC) \cite{hu2019doubly}, and one-pass incomplete multi-view clustering (OPIMC) \cite{shao2016online}. Besides these, graph regularized partial multi-view clustering (GPMVC) \cite{rai2016partial} proposed by Rai et al. can also be regarded as a variant of the weighted MF based IMC method.

In this paper, from the viewpoint of the strategy of achieving the consensus representation, we can unify the existing weighted MF based IMC methods into the following two frameworks, referred to as WMF\_IMCF1 and WMF\_IMCF2:
\begin{equation}\label{eq_6}\small
\text{WMF\_IMCF1:}\begin{aligned}
\mathop {\min }\limits_\Upsilon  &\sum\limits_{v = 1}^l \begin{array}{l}
\left\| {\left( {{Y^{\left( v \right)}} - {U^{\left( v \right)}}{\bar P^{\left( v \right)}}} \right){W^{\left( v \right)}}} \right\|_F^2\\
 + {\gamma _v}\left\| {\left( {{Q^{\left( v \right)}}{\bar P^{\left( v \right)}} - P} \right){W^{\left( v \right)}}} \right\|_F^2
\end{array}  \\
&+ \varphi \left( {{U^{\left( v \right)}},{\bar P^{\left( v \right)}},P} \right) \\
&s.t.{\kern 1pt} {\kern 1pt} \left\{ {P,\left\{ {{U^{\left( v \right)}},{\bar P^{\left( v \right)}}} \right\}_{v = 1}^l} \right\} \in \phi
\end{aligned}
\end{equation}
\begin{equation}\label{eq_7}\small
\text{WMF\_IMCF2:}\begin{aligned}
\mathop {\min }\limits_{P,\left\{ {{U^{\left( v \right)}}} \right\}_{v = 1}^l} &\sum\limits_{v = 1}^l {\left\| {\left( {{Y^{\left( v \right)}} - {U^{\left( v \right)}}P} \right){W^{\left( v \right)}}} \right\|_F^2}  \\
&+ \varphi \left( {{U^{\left( v \right)}},P} \right) \\
&s.t.{\kern 1pt} {\kern 1pt} \left\{ {P,\left\{ {{U^{\left( v \right)}}} \right\}_{v = 1}^l} \right\} \in \phi
\end{aligned}
\end{equation}
where $\Upsilon  = \left\{ {P,\left\{ {{U^{\left( v \right)}},{\bar P^{\left( v \right)}}} \right\}_{v = 1}^l} \right\}$. ${U^{\left( v \right)}}$ and ${\bar P^{\left( v \right)}} \in {R^{c \times n}}$ are the basis matrix and latent representation of all instances (\emph{i.e.}, including the available instances and missing instances) of the $v$th view. ${Q^{\left( v \right)}} \in {R^{c \times c}}$ is a diagonal weight matrix, which is defined as $I$ in MIC and OMVC, and defined as $Q_{i,i}^{\left( v \right)} = \sum\limits_{j = 1}^{{m_v}} {U_{j,i}^{\left( v \right)}} $ in GPMVC{\footnote{When 1) the diagonal weighted matrix ${W^{\left( v \right)}}$ is defined as $W_{i,i}^{\left( v \right)} = 1$ for the $i$th sample that has the $v$th view otherwise $W_{i,i}^{\left( v \right)} = 0$. 2) $\varphi \left( {{U^{\left( v \right)}},{\bar P^{\left( v \right)}},P} \right) = \sum\limits_{v = 1}^l {{\beta _v}Tr\left( {{\bar P^{\left( v \right)}}{G^{\left( v \right)}}{L_{Z^{\left( v \right)}}}{G^{\left( v \right)T}}{\bar P^{\left( v \right)T}}} \right)} $, where ${G^{\left( v \right)}}$ is defined in Table V at the supplementary file, ${L_{Z^{\left( v \right)}}}$ is the Laplacian matrix constructed from the available instances of the $v$th view. 3) $\phi \left( {P,\left\{ {{U^{\left( v \right)}},{\bar P^{\left( v \right)}}} \right\}_{v = 1}^l} \right) = \left\{ {\left\{ {{U^{\left( v \right)}},{\bar P^{\left( v \right)}}} \right\}_{v = 1}^l \ge 0} \right\}$, then framework (\ref{eq_6}) is equivalent to the learning model of GPMVC \cite{rai2016partial}.}} . ${W^{\left( v \right)}}$ is a diagonal weight matrix to reduce the negative influence of the missing views. In the above two frameworks, $\phi$ denotes the boundary constraint, such as the non-negative constraint in MIC and OMVC. $\varphi \left( {{U^{\left( v \right)}},{\bar P^{\left( v \right)}},{P}} \right)$ and $\varphi \left( {{U^{\left( v \right)}},P} \right)$ denote the regularization constraint of these variables.

\textbf{Analysis of the two frameworks}: MIC, OMVC, and GPMVC are the most representative works of WMF\_IMCF1. The most representative methods of WMF\_IMCF2 are DAIMC \cite{hu2019doubly} and OPIMC \cite{hu2019one}. From models (\ref{eq_6}) and (\ref{eq_7}), we can find that WMF\_IMCF1 obtains the consensus representation from the latent representations derived from all views, while WMF\_IMCF2 directly decomposes the original multi-view data into a consensus representation and several basis matrices. Intuitively, compared with WMF\_IMCF2, WMF\_IMCF1 provides more freedom in learning the consensus representation. However, WMF\_IMCF1 introduces at least an extra tunable hyper-parameter ${\gamma _v}$, which increases the complexity of the optimal parameter selection.

\textbf{Connections of the existing weighted MF based methods}: From Fig.7, we can observe that these weighted MF methods commonly impose different constraints on the basis matrix or consensus representation for capturing different properties. Among these methods, MIC, OMVC, and OPIMC impose the ${l_{2,1}}$, ${l_1}$, and ${l_F}$ norm based constraints on the latent representations or basis matrices, respectively. These approaches can avoid the trivial solutions, but are not beneficial to improve the discriminability. Compared with these methods, DAIMC and GPMVC attempt to explore more information from the data by imposing the approximate orthogonal constraint or graph constraint, which are beneficial to achieve a better performance. Nevertheless, OMVC and OPIMC have their superiorities in efficiency and memory cost since their objective functions can be optimized in a chunk by chunk style. In addition to the above connections, there are many differences among these methods. For example, \emph{1) Filling of the missing instance (Definition of ${Y^{\left( v \right)}} \in {R^{{m_v} \times n}}$)}. Some methods, such as MIC and OMVC, use the average instance to fill in the missing instances. The other methods, such as DAIMC and OPIMC, set all elements of the missing instances as 0. \emph{2) Definition of the diagonal weighted matrices $\{{W^{\left( v \right)}}\}_{v=1}^l$}. Generally, the diagonal element $W_{i,i}^{\left( v \right)}$ is set as 1 if the $i$th sample has the $v$th view. However, if the $v$th view is missing for the $i$th sample, $W_{i,i}^{\left( v \right)}$ is set as ${{{n_v}} \mathord{\left/
 {\vphantom {{{n_v}} n}} \right.
 \kern-\nulldelimiterspace} n}$ in MIC and 0 in DAIMC. It should be noted that when $W_{i,i}^{\left( v \right)}$ is set as ${{{n_v}} \mathord{\left/
 {\vphantom {{{n_v}} n}} \right.
 \kern-\nulldelimiterspace} n}$ for the missing instances, the filled virtual instances will affect the consensus representation learning.
\begin{figure}[ht]
\centering
\includegraphics[width=3.5in,height=2in]{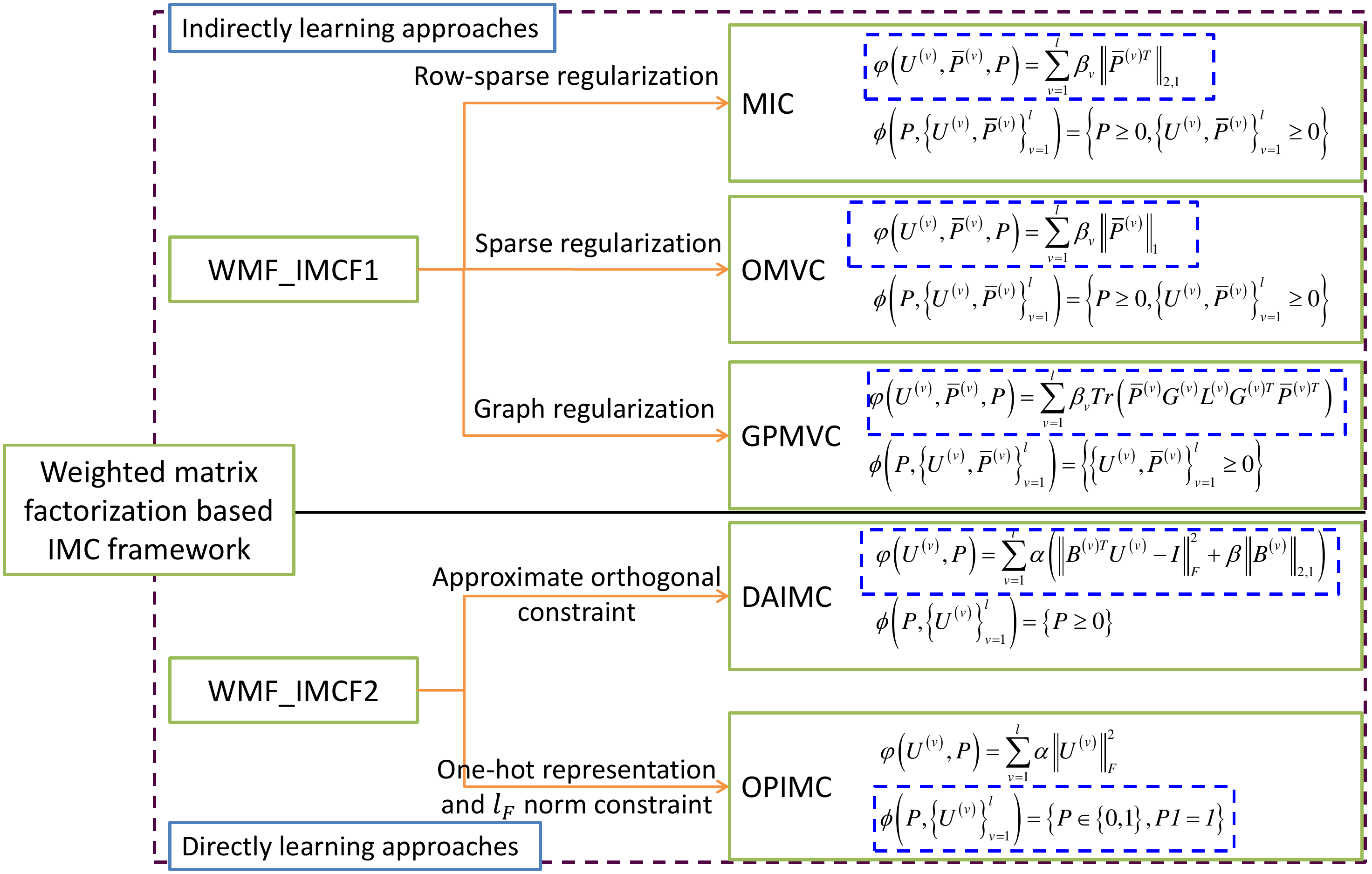}
\caption{Frameworks of the weighted MF based IMC.}
\label{fig:fig5}
\end{figure}

Compared with the models of PMVCF, WMF\_IMCF1, WMF\_IMCF2, and the conventional MF based multi-view clustering model (1), we can find that PMVCF, WMF\_IMCF1, and WMF\_IMCF2 generally focus on exploring the partially aligned information of the available instances to address the incomplete learning problem, where PMVCF designs the partial MF model and the other two frameworks introduce the weighted constraint. However, these methods suffer from the following problems: 1) they ignore the hidden information of missing views; 2) these methods are all based on the squared Euclidean distance or Frobenius norm, which may be not powerful enough to capture the intrinsic structure of the incomplete multi-view data. To address the first issue, two improved methods are proposed from the viewpoint of missing view recovery. For example, in \cite{xu2015multi}, Xu et al. proposed a very simple missing-view restoration and common-representation joint learning model based on the assumption that all views are generated from a shared subspace. Besides this, Wen et al. proposed a missing-view restoration method, called unified embedding alignment framework (UEAF) \cite{wen2019unified}. UEAF can be viewed as an improved version of the method proposed in \cite{xu2015multi} whose learning model is exactly the first item of UEAF. Compared with the method proposed in \cite{xu2015multi}, UEAF further considers the geometric structure of data and features, and thus has the potential to perform better on missing-view restoration and clustering. However, a shortcoming of UEAF is that it has a relatively high computation complexity. For the second issue, kmeans-based consensus IMC methods proposed in \cite{liu2018consensus} and \cite{wu2014k} provide some new ideas, where different distance metrics, such as KL-divergence and Cosine distance, are unified into their proposed frameworks. The two methods can be viewed as the variants of multi-view kmeans. Moreover, a common point between the two methods and the previous MF based frameworks is that they adopt a similar weighted strategy as in WMF\_IMCF1 and WMF\_IMCF2 to reduce the negative influence of missing views. More specifically, when choosing the square Euclidean distance to measure the matrix decomposition errors, the learning model proposed in \cite{liu2018consensus} can be unified into WFM\_IMCF1.
\section{Kernel learning based IMC}
Multiple kernel clustering (MKC) generally seeks to learn a consensus representation or multiple latent representations corresponding to all views from the multiple kernels pre-constructed from all views, followed by kmeans to achieve the clustering results. The conventional MKCs all require that the input kernels are complete. In other words, the existing conventional MKCs cannot handle the kernel clustering tasks where some rows and columns of the pre-constructed kernels are absent, caused by the missing views. To solve this problem, many researchers have studied the multiple incomplete kernel clustering (MIKC) in the recent decade.

Most MIKCs solve the incomplete learning problem based on recovering the missing rows and columns of the kernel matrices. In view of the exploited main techniques to solve the incomplete issue, we divide the existing MIKCs into two groups, where the first group is based on the Laplacian regularization and kernel canonical correlation analysis (KCCA), and the second group is based on multiple kernel kmeans.
\subsection{Laplacian regularization and KCCA based IMC}
Laplacian regularization and KCCA based IMC methods generally first perform the kernel matrices completion and then learn the latent representations of all views via KCCA. For example, based on a complete kernel pre-constructed from the view without missing instances, Trivedi et al. recovered the missing elements of the incomplete kernel by solving the following Laplacian regularized problem \cite{trivedi2010multiview}:

\begin{equation}\label{eq_8}
\begin{array}{l}
\mathop {\min }\limits_{{K^{\left( 2 \right)}} \ge 0} Tr\left( {{L^{\left( 1 \right)}}{K^{\left( 2 \right)}}} \right){\kern 1pt} \\
{\kern 1pt} s.t.{\kern 1pt} {\kern 1pt} K_{i,j}^{\left( 2 \right)} = k\left( {x_i^{\left( 2 \right)},x_j^{\left( 2 \right)}} \right),\forall i,j = 1,2, \ldots ,{n_2}
\end{array}
\end{equation}
where ${L^{\left( 1 \right)}}$ denotes the Laplacian matrix of kernel ${K^{\left( 1 \right)}} \in {R^{n \times n}}$ of the $\emph{1}$st view without missing instances. $k(x_i^{(2)},x_j^{(2)})$ is a value computed by a kernel function $k(*,*)$.

Then, Trivedi et al. performed KCCA on the complete kernel ${K^{\left( 1 \right)}}$ and recovered complete kernel ${K^{\left( 2 \right)}}$ to obtain the latent representations of two views, followed by kmeans clustering. In this paper, we refer to the method proposed by Trivedi et al. as multiple incomplete kernel clustering with one complete kernel (MIKC\_OCK).

To solve the complete view issue of MIKC\_OCK, collective kernel learning (CoKL) is proposed, which interactively recovers the incomplete kernel matrices by optimizing a similar problem as (\ref{eq_8}) across different views \cite{shao2013clustering}. One of the limitations of CoKL is that it is only suitable to the incomplete data with two views.

In summary, the Laplacian regularization and KCCA based IMC methods are not suitable to practical applications since these methods can only handle one kind of incomplete cases. Moreover, the two-step based approach, \emph{i.e.}, perform the kernel matrices completion and latent representations learning separately, cannot guarantee the global optimal kernel matrices and latent representations.
\subsection{Multiple kernel kmeans based IMC}
Compared with the Laplacian regularization and KCCA based IMC methods, multiple kernel kmeans based IMC methods seek to recover the kernel matrices and learn the consensus representation or cluster index matrix simultaneously in a joint framework. One of the basic and intuitive multiple incomplete kernel kmeans clustering models proposed in \cite{liu2017multiple} is expressed as follows:
\begin{equation}\label{eq_9}\small
\begin{array}{l}
\mathop {\min }\limits_{\left\{ {{\alpha ^{\left( v \right)}},{K^{\left( v \right)}}} \right\}_{v = 1}^l,P} Tr\left( {\sum\limits_{v = 1}^l {{{\left( {{\alpha ^{\left( v \right)}}} \right)}^2}{K^{\left( v \right)}}} \left( {I - P{P^T}} \right)} \right)\\
s.t.{\kern 1pt} {\kern 1pt} {\kern 1pt} {P^T}P = I,\sum\limits_{v = 1}^l {{\alpha ^{\left( v \right)}}}  = 1,{\alpha ^{\left( v \right)}} \ge 0,K_{\left( {{s_v},{s_v}} \right)}^{\left( v \right)} = K_{SV}^{\left( v \right)},{K^{\left( v \right)}} \ge 0
\end{array}
\end{equation}
where $P \in {R^{n \times c}}$ denotes the consensus representation. ${s_v}$ ($1 \le p \le {n_v}$) denotes the sample indexes of those instances that are observed in the $v$th view, $K_{SV}^{\left( v \right)}$ denotes the sub-kernel matrix pre-constructed from these available instances. ${K^{\left( v \right)}} \in {R^{n \times n}}$ is the kernel matrix of the $v$th view to recover. $K_{\left( {{s_v},{s_v}} \right)}^{\left( v \right)}$ denotes the elements corresponding to the observed instances. 

For convenience, we refer to the method proposed in \cite{liu2017multiple} as incomplete multiple kernel kmeans clustering (IMKKC). From (\ref{eq_9}), we can observe that the main idea of IMKKC is to align the fused kernel $\sum\limits_{v = 1}^l {{{\left( {{\alpha ^{\left( v \right)}}} \right)}^2}{K^{\left( v \right)}}} $ and ideal kernel $P{P^T}$. Different from IMKKC, another similar basic model, referred to as consensus kernel kmeans based IMC (CKK-IMC) proposed in \cite{ye2017consensus} seeks to obtain the consensus $P$ from the latent representations $\left\{ {{P^{\left( v \right)}} \in {R^{n \times c}}} \right\}_{v = 1}^l$ by introducing the dissimilarity based regularization item. The main issues of IMKKC and CKK-IMC are that the two methods ignore the local structure of the data and do not sufficiently consider the complementary information of views. Besides these, IMKKC has a relatively high computational complexity of $O\left( {{n^3}} \right)$. Based on IMKKC, many improved methods, such as localized incomplete multiple kernel kmeans clustering (LIMKKC) \cite{zhu2018localized} and incomplete multiple kernel kmeans with mutual kernel completion (MKKM-IK-MKC) \cite{liu2019multiple} are proposed. In particular, LIMKKC mainly introduces some neighborhood indication matrices to IMKKC to preserve the local information of the data. In order to recover the missing rows and columns of the kernel matrices better, MKKM-IK-MKC introduces a sparse reconstruction based constraint to model (\ref{eq_9}) to capture more complementary information from the kernels.

Generally, the above kernel completion based methods need to recover $\frac{1}{2}\left( {n - {n_v}} \right)\left( {n + {n_v} + 1} \right)$ elements for the $v$th view with ${n_v}$ observed instances. However, recovering such a large number of elements may trap the model into a local minimum, which affects the clustering performance \cite{liu2019efficient}. In \cite{liu2019efficient}, Liu et al. proposed another multiple kernel based approach, named efficient and effective incomplete multi-view clustering (EE-IMVC), to address the IMC problem. Different from the above multiple kernel kmeans based methods, EE-IMVC does not focus on recovering the kernel matrices. It seeks to jointly compute the latent representations corresponding to the missing instances of each view and obtain the consensus representation $P \in {R^{n \times c}}$ as follows:
\begin{equation}\label{eq_10}
\begin{array}{l}
\mathop {\max }\limits_\Upsilon  Tr\left( {{P^T}\sum\limits_{v = 1}^l {{\alpha ^{\left( v \right)}}\left[ \begin{array}{l}
\bar P_{\left( o \right)}^{\left( v \right)}\\
\bar P_{\left( u \right)}^{\left( v \right)}
\end{array} \right]{Q^{\left( v \right)}}} } \right)\\
s.t.{\kern 1pt} {\kern 1pt} {P^T}P = I,{Q^{\left( v \right)T}}{Q^{\left( v \right)}} = I,\\
\bar P_{\left( u \right)}^{\left( v \right)T}\bar P_{\left( u \right)}^{\left( v \right)} = I,\sum\limits_{v = 1}^l {{{\left( {{\alpha ^{\left( v \right)}}} \right)}^2}}  = 1
\end{array}
\end{equation}
where $\Upsilon  = \left\{ {P,\left\{ {{Q^{\left( v \right)}},\bar P_{\left( u \right)}^{\left( v \right)},{\alpha ^{\left( v \right)}}} \right\}_{v = 1}^l} \right\}$. Permutation matrix ${Q^{\left( v \right)}} \in {R^{c \times c}}$ is used to match ${\bar P^{\left( v \right)}} = \left[ \begin{array}{l}
\bar P_{\left( o \right)}^{\left( v \right)}\\
\bar P_{\left( u \right)}^{\left( v \right)}
\end{array} \right]$ and $P$. $\bar P_{\left( o \right)}^{\left( v \right)}$ and $\bar P_{\left( u \right)}^{\left( v \right)}$ are the latent representations of the observed instances and missing instances of the $v$th view, respectively.

Compared with the other multiple kernel kmeans based methods, EE-IMVC greatly reduces the computational complexity and memory cost. However, it cannot bring the nearest sample pairs closer and push the other sample pairs far away.
\section{Graph learning based IMC}
Similar to the conventional multi-view spectral clustering, the purpose of graph learning based IMC is to obtain a consensus graph or consensus representation from multiple incomplete graphs constructed from the data, where almost all of the existing methods are based on the pre-constructed incomplete graphs and the missing similarity elements are set as 0 or average values. In particular, we can divide the existing methods into three categories as shown in Fig.8.
\begin{figure*}[!htb]
\centering
\subfloat[]{\includegraphics[width=2.2in,height=1in]{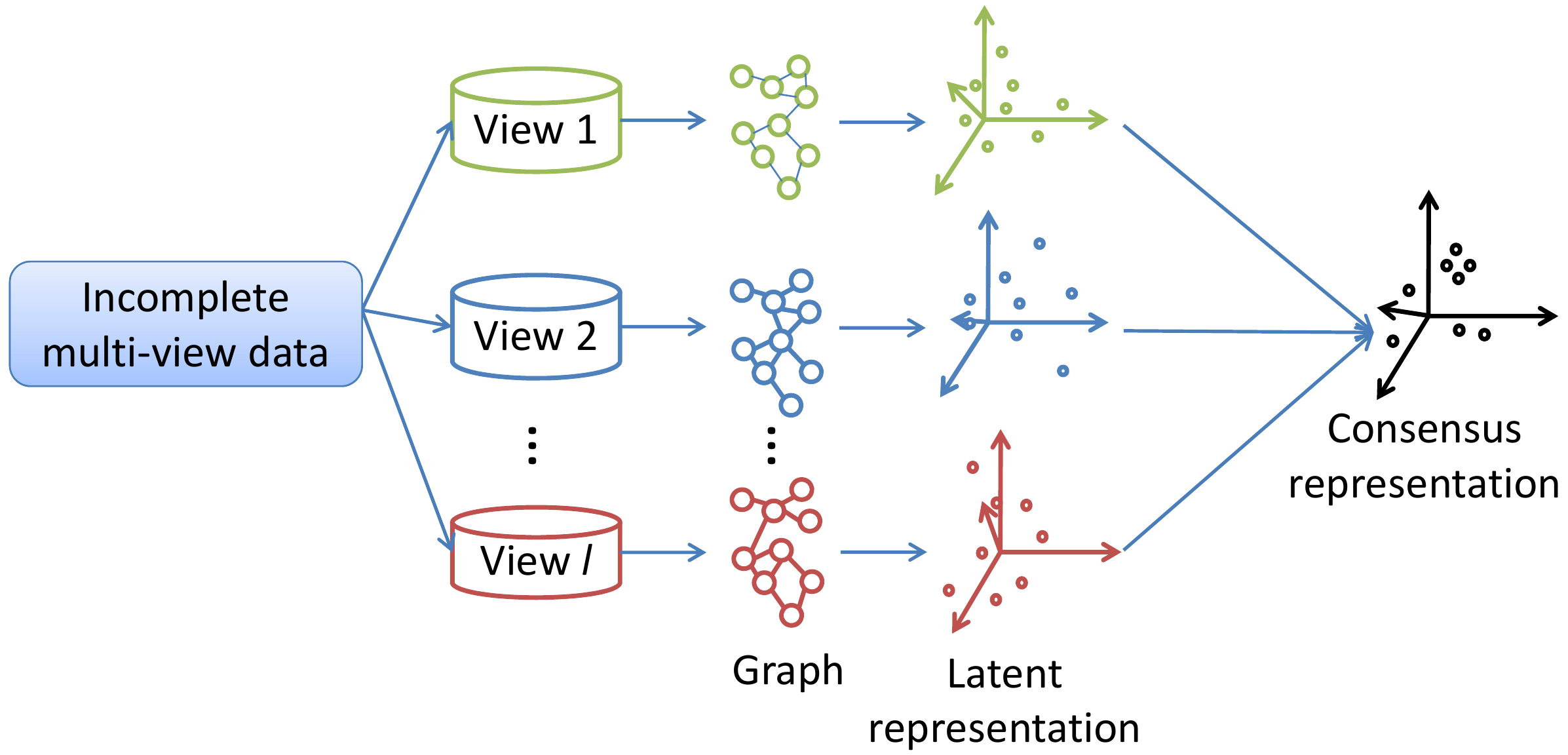}}
\hfil
\subfloat[]{\includegraphics[width=2.2in,height=1in]{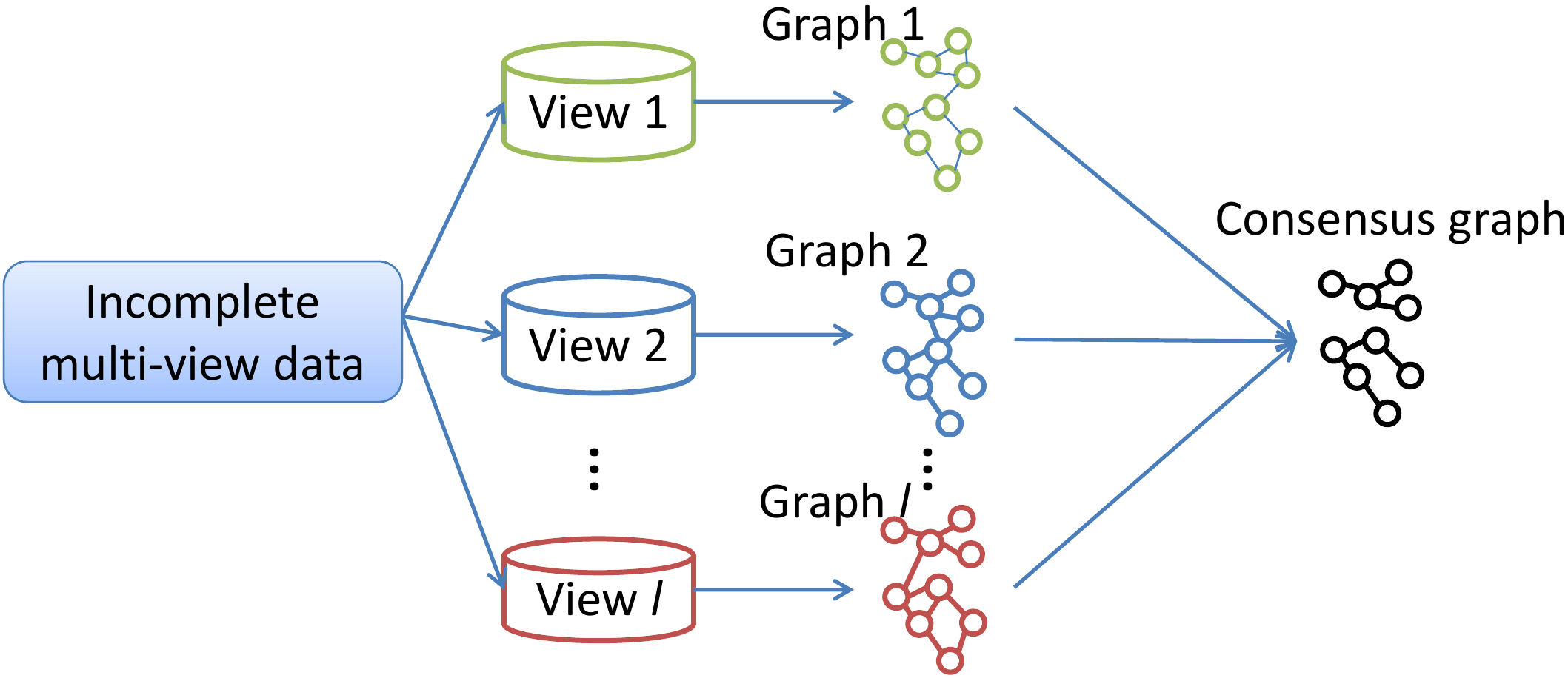}}
\hfil
\subfloat[]{\includegraphics[width=2.2in,height=1in]{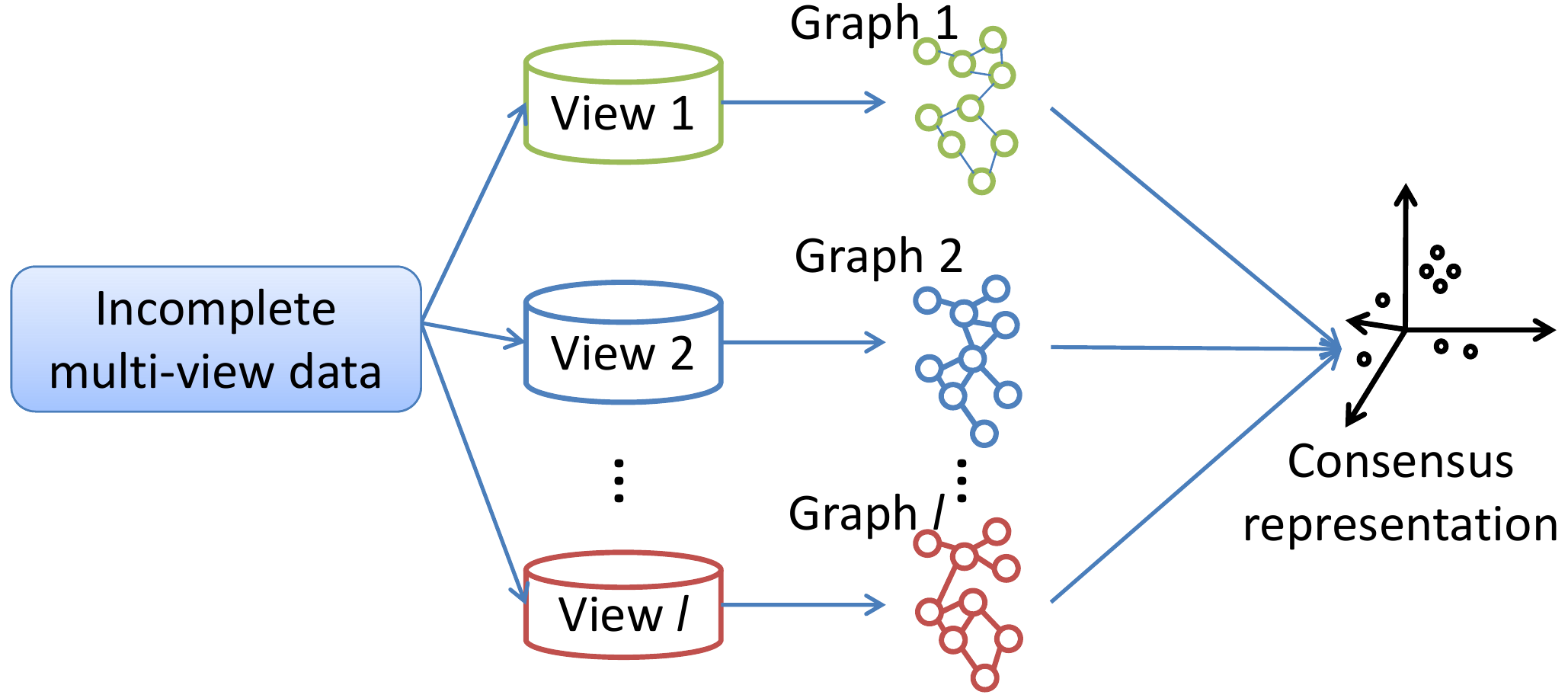}}
\caption{Three kinds of graph learning based IMC methods, where (a) and (c) focus on obtaining the consensus representation, and (b) seeks to obtain the consensus graph from the incomplete data.}
\end{figure*}

As a representative method of Fig.8(a), spectral clustering based IMC (SCIMC) exploits a very simple co-training approach to recover the latent representations of missing instances and obtain the consensus representation \cite{gao2016incomplete}. Given the similarity matrices $\left\{ {{Z^{\left( v \right)}}} \right\}_{v = 1}^l$ with a size of $n \times n$ for each view, where missing entries are filled in the average of the column vectors corresponding to the observed instances, SCIMC seeks to obtain the consensus representation $P \in {R^{n \times c}}$ by optimizing the following four problems one by one:
\begin{equation}\label{eq_11}\small
\left\{ \begin{array}{l}
\mathop {\max }\limits_{{\bar P^{\left( v \right)T}}{\bar P^{\left( v \right)}} = I} Tr\left( {{\bar P^{\left( v \right)T}}{L^{\left( v \right)}}{\bar P^{\left( v \right)}}} \right)\\
\mathop {\max }\limits_{{P^T}P = I} \sum\limits_{v = 1}^l {{\lambda _v}Tr\left( {P{P^T}{\bar P^{\left( v \right)}}{\bar P^{\left( v \right)T}}} \right)} \\
\mathop {\max }\limits_{\bar P_m^{\left( v \right)}} Tr\left( {\left[ \begin{array}{l}
\bar P_a^{\left( v \right)}\\
\bar P_m^{\left( v \right)}
\end{array} \right]{{\left[ \begin{array}{l}
\bar P_a^{\left( v \right)}\\
\bar P_m^{\left( v \right)}
\end{array} \right]}^T}\left[ \begin{array}{l}
{P_a}\\
{P_m}
\end{array} \right]{{\left[ \begin{array}{l}
{P_a}\\
{P_m}
\end{array} \right]}^T}} \right)\\
\mathop {\max }\limits_{{P^T}P = I} \sum\limits_{v = 1}^l {{\lambda _v}Tr\left( {P{P^T}{\bar P^{\left( v \right)}}{\bar P^{\left( v \right)T}}} \right)}
\end{array} \right.
\end{equation}
where ${L^{\left( v \right)}}$ is a symmetrical normalized matrix calculated as ${L^{\left( v \right)}} = {\left( {{D^{\left( v \right)}}} \right)^{{{ - 1} \mathord{\left/
 {\vphantom {{ - 1} 2}} \right.
 \kern-\nulldelimiterspace} 2}}}{Z^{\left( v \right)}}{\left( {{D^{\left( v \right)}}} \right)^{{{ - 1} \mathord{\left/
 {\vphantom {{ - 1} 2}} \right.
 \kern-\nulldelimiterspace} 2}}}$ and diagonal matrix $D_{i,i}^{\left( v \right)} = \sum\limits_{j = 1}^n {Z_{i,j}^{\left( v \right)}} $. In the third step, $P$ and ${\bar P^{\left( v \right)}}$ are reordered as $\left[ \begin{array}{l}
{P_a}\\
{P_m}
\end{array} \right]$ and $\left[ \begin{array}{l}
\bar P_a^{\left( v \right)}\\
\bar P_m^{\left( v \right)}
\end{array} \right]$, where $\bar P_a^{\left( v \right)}$ and $\bar P_m^{\left( v \right)}$ denote the latent representation of the available instances and missing instances of the $v$th view, respectively.

From (\ref{eq_11}) and the conventional multi-view spectral clustering model (2), we can find that SCIMC divides the conventional model (2) into several independent steps and seeks to alternately recover the latent representation corresponding to the missing views by borrowing the information of the other views. Then it combines the latent representations of all views to achieve the consensus representation by minimizing the disagreement of different views. In (\ref{eq_11}), the first two steps can be viewed as the initialization steps. Although SCIMC can handle the IMC problem, it suffers from three issues: 1) It cannot achieve the optimal consensus representation by optimizing the four problems independently. 2) It is unreasonable to initialize the similarity graphs by setting the missing entries as the average of the columns. 3) SCIMC is sensitive to the pre-constructed similarity graph $\left\{ {{Z^{\left( v \right)}}} \right\}_{v = 1}^l$.

Wang et al. proposed a perturbation-oriented incomplete multi-view clustering (PIC) method based on the spectral perturbation theory, which is a representative work of Fig.8 (b) \cite{wang2019spectral}. For any incomplete data, PIC mainly adopts the following three steps to obtain the consensus graph for spectral clustering. \emph{1) Nearest-neighbor graph construction and completion.} PIC provides a graph learning method to directly obtain some nearest-neighbor graphs $\left\{ {{Z^{\left( v \right)}} \in {R^{n \times n}}} \right\}_{v = 1}^l$, where some rows of ${Z^{\left( v \right)}}$ associated to the missing instances are set as the average of the rows corresponding to the other available views. \emph{2) Fusion weight calculation}. PIC establishes a perturbation-oriented model to learn some coefficients for graph fusion. \emph{3) Consensus graph learning}. PIC obtains the consensus graph by fusing these multiple graphs with the learned coefficients. PIC can handle all kinds of incomplete multi-view data. However, since the graph construction, weight calculation, and consensus graph learning are independent to each other, PIC is also sensitive to the quality of the pre-constructed graphs.

Recently, Zhou et al. proposed a consensus graph learning approach for IMC (CGL\_IMC) \cite{zhou2019consensus}. Different from PIC, CGL\_IMC provides a weighted graph learning approach to construct the similarity graphs $\left\{ {{Z^{\left( v \right)}} \in {R^{n \times n}}} \right\}_{v = 1}^l$ of all incomplete views. Then, it obtains the consensus graph from the following Laplacian regularized graph learning framework:

\begin{equation}\label{eq_12}\small
\begin{array}{l}
\mathop {\min }\limits_{{Z^*},{\alpha ^{\left( v \right)}}} \sum\limits_{v = 1}^l {{\alpha ^{\left( v \right)}}\left\| {{Z^*} - {Z^{\left( v \right)}}} \right\|_F^2}  + 2\beta Tr\left( {{F^T}{L_{{Z^*}}}F} \right)\\
s.t.{\kern 1pt} {\kern 1pt} {\alpha ^{\left( v \right)}} \ge 0,\sum\limits_{v = 1}^l {{\alpha ^{\left( v \right)}}}  = 1,{F^T}F = I,{Z^*} \ge 0,{Z^{*T}}\emph{1} = \emph{1}
\end{array}
\end{equation}
where ${Z^*} \in {R^{n \times n}}$ denotes the consensus graph.

By introducing the variant of rank constraint $Tr\left( {{F^T}{L_{{Z^*}}}F} \right)$, CGL\_IMC has the potential to learn the optimal consensus graph that has exactly $c$ blocks for the data with $c$ clusters. However, similar to PIC, CGL\_IMC is also sensitive to the quality of the pre-constructed graphs.

In \cite{wen2018incompleteTCYB}, Wen et al. proposed a spectral clustering based method, named incomplete multi-view spectral clustering with adaptive graph learning (IMSC\_AGL). As shown in Fig.8(c), IMSC\_AGL tries to directly obtain the consensus representation from multiple graphs adaptively learned from the incomplete data. In particular, different from SCIMC and PIC, IMSC\_AGL integrates the adaptive graph construction and spectral based consensus representation learning into a joint optimization framework, which can naturally address the issues of PIC and SCIMC. The learning model of IMSC\_AGL is as follows:
\begin{equation}\label{eq_13}
\begin{array}{l}
\mathop {\min }\limits_{{Z^{\left( v \right)}},{E^{\left( v \right)}},P} \sum\limits_{v = 1}^l {\left( {{{\left\| {{Z^{\left( v \right)}}} \right\|}_*} + {\lambda _2}{{\left\| {{E^{\left( v \right)}}} \right\|}_1}} \right)} \\
 + {\lambda _1}\sum\limits_{v = 1}^l {Tr\left( {P{G^{\left( v \right)}}L_{Z^{(v)}}{G^{\left( v \right)T}}{P^T}} \right)} \\
s.t.{\kern 1pt} {\kern 1pt} {X^{\left( v \right)}} = {X^{\left( v \right)}}{Z^{\left( v \right)}} + {E^{\left( v \right)}},{Z^{\left( v \right)}}\emph{1} = \emph{1},\\
0 \le {Z^{\left( v \right)}} \le 1,Z_{i,i}^{\left( v \right)} = 0,P{P^T} = I
\end{array}
\end{equation}
where $P \in {R^{c \times n}}$ is the consensus representation. $L_{Z^{(v)}}$ is the Laplacian matrix of graph ${Z^{\left( v \right)}}$. ${G^{\left( v \right)}}$ is defined in Table V at the supplementary file.

From the second item of the learning model (\ref{eq_13}), we can observe that IMSC\_AGL does not introduce any uncertain information to guide the consensus representation learning like SCIMC and PIC. This ensures IMSC\_AGL obtain a more reasonable consensus representation. However, IMSC\_AGL suffers from the issue of high computational complexity, which is not suitable for large-scale datasets.

All of the above graph learning based methods commonly separate the model optimization and clustering into two independent steps, which need to implement kmeans as post-processing to partition data into respective groups. In \cite{zhuge2019simultaneous}, Zhuge et al. proposed a unified framework, named simultaneous representation learning and clustering (SRLC), which incorporates the graph based representation learning and label prediction into a joint framework. Compared with the previous works, SRLC has the potential to obtain the global optimal clustering labels. However, its performance is also sensitive to the quality of graphs pre-constructed from data.

Generally speaking, compared with MF based methods, graph learning based methods can better excavate the geometric information of the data. However, since the graph learning based methods need to implement some relatively time-consuming operations, such as eigenvalue computation, singular value decomposition, and matrix inverse operation, these methods may not be suitable for large-scale datasets. Thus, it is necessary to develop some efficient graph learning algorithms for IMC. In addition, in view of the fact that there are a lot of information residing in the data, such as the global information and local information, it is better to sufficiently consider these valuable information in IMC models.
\section{Deep learning based IMC}
As presented in the previous section, representation learning plays an important role in most of the existing IMC methods. Learning a more discriminative consensus representation is crucial to achieve a better performance. In recent years, deep learning has been successfully applied in many fields of computer vision and pattern classification owing to its good performance in learning a high-level feature representation. To this end, researchers seek to combine the deep learning and conventional IMC approach to improve the performance, where the most representative works are incomplete multi-view clustering via deep semantic mapping (IMC\_DSM) \cite{zhao2018incomplete}, partial multi-view clustering via consistent generative adversarial networks (PMVC\_CGAN) \cite{wang2018partial}, and adversarial incomplete multi-view clustering (AIMC) \cite{xu2019adversarial}.

IMC\_DSM integrates the deep neural networks (DNN) based feature extraction, PMVC, and local graph regularization into a framework as shown in Fig.9. The objective function of IMC\_DSM is expressed as follows:
\begin{equation}\label{eq_14}
\begin{array}{l}
\mathop {\min }\limits_{{P_c},\left\{ {{U^{\left( v \right)}},{P_s^{\left( v \right)}}} \right\}_{v = 1}^l} \sum\limits_{v = 1}^l \begin{array}{l}
\left\| {\left[ {A_c^{\left( v \right)},{A_s^{\left( v \right)}}} \right] - {U^{\left( v \right)}}\left[ {{P_c},{P_s^{\left( v \right)}}} \right]} \right\|_F^2\\
 + {\lambda _v}Tr\left( {{P^{\left( v \right)}}{L_{Z^{\left( v \right)}}}{P^{\left( v \right)T}}} \right)
\end{array} \\
s.t.{\kern 1pt} {\kern 1pt} {U^{\left( v \right)}} \ge 0,{P^{\left( v \right)}} = \left[ {{P_c},{P_s^{\left( v \right)}}} \right] \ge 0
\end{array}
\end{equation}
where ${A^{\left( v \right)}} = \left[ {A_c^{\left( v \right)},{A_s^{\left( v \right)}}} \right] = f\left( {{\vartheta^{\left( v \right)}}{X^{\left( v \right)}} + {b^{\left( v \right)}}} \right)$ denotes the feature representation produced by DNN in the $v$th view. ${L_{Z^{\left( v \right)}}}$ is the Laplacian matrix of the nearest-neighbor graph pre-constructed from the available instances of the $v$th view. $\vartheta^{(v)}$ and $b^{(v)}$ are the network weights and bias, respectively.
\begin{figure}[ht]
\centering
\includegraphics[width=3.3in,height=1.2in]{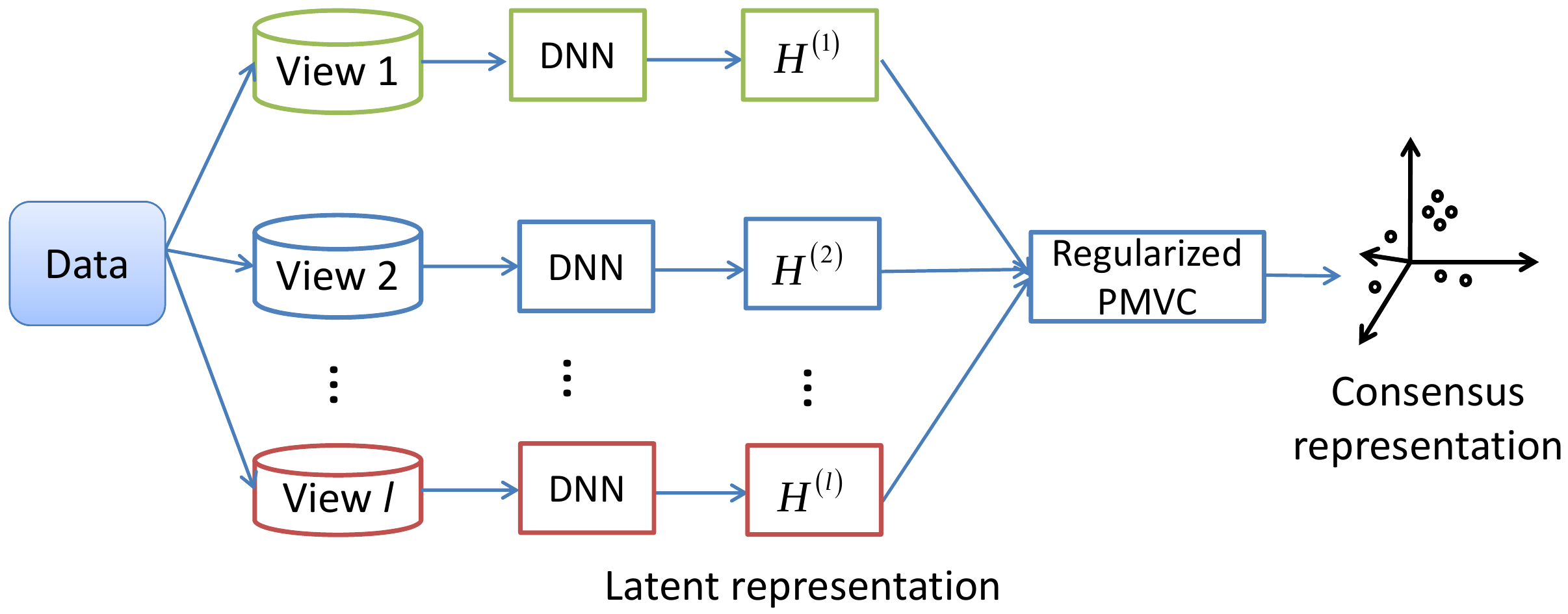}
\caption{Framework of IMC\_DSM. For simplicity, we refer to the objective function (13) as regularized PMVC.}
\label{fig:fig5}
\end{figure}

It is easy to observe that: 1) the objective function (\ref{eq_14}) is a special case of our unified framework (5). 2) IMC\_DSM addresses the incomplete learning problem by exploring the partially aligned information among the available views as the conventional PMVCF introduced in the previous section. In IMC\_DSM, DNN can extract the high-level features from the data and the local graph regularization item is beneficial to obtain a more reasonable structured representation.

PMVC\_CGAN provides another clustering approach for incomplete two-view data based on autoencoder (AE) and GAN. PMVC\_CGAN mainly contains three components: 1) It exploits AE to generate the latent representations of all views. 2) GAN is introduced to generate the missing views. 3) A `KL-divergence' based loss function is introduced to guarantee that the learned latent representation is suitable for the clustering task. In particular, PMVC\_CGAN provides many interesting points in comparison with the existing works, including IMC\_DSM. For example, existing methods need to implement kmeans on the obtained representation for achieving the clustering result, while PMVC\_CGAN can directly produce the final clustering result according to the `KL-divergence'. In addition, almost all of the existing works ignore the information of missing views, while PMVC\_CGAN can sufficiently exploit it for model training via GAN. Moreover, the most difference between PMVC\_CGAN and the existing works is that PMVC\_CGAN focuses on clustering, while the other works focus on learning the consensus representation or graph.
To address the issue that PMVC\_CGAN cannot handle the incomplete data with more than two views, Xu et al. proposed AIMC. AIMC can be viewed as an extension of PMVC\_CGAN for the case of more than two views since it exploits a similar framework as PMVC\_CGAN. However, AIMC requires that some samples must have complete views.

From the above presentations, we can observe that: 1) exploiting the deep neural network is beneficial to learn more discriminative representations so as to improve the clustering performance. 2) The deep based methods can be applied to large-scale datasets owing to the batch based training style. 3) The biggest shortcoming of the existing deep based methods is that these deep based methods cannot be applied to all kinds of incomplete cases.

In Table VI of the supplementary material, we summarize the main merits and drawbacks of some representative IMC methods mentioned above (refer to Section III-VI). From the presentation of these methods, we can obtain the following two points: 1) Imposing some weighted matrices constructed from the view-missing information on the conventional multi-view clustering methods is an effective and flexible approach to address the incomplete learning problem. 2) Designing some robust models to recover the missing views or missing elements in the kernels or graphs is a valuable research direction to address the IMC problem. In the second approach, a challenging problem is how to ensure the rationality of missing-view or missing-element recovery theoretically.
\section{Experiments}
In this section, we mainly conduct several experiments to compare and analyze some representative IMC methods presented in the previous sections. The compared methods include best single view (BSV) \cite{zhao2016incomplete}, Concat \cite{zhao2016incomplete}, multi-view non-negative MF (MultiNMF) \cite{liu2013multi}, auto-weighted multiple graph learning (AMGL) \cite{nie2016parameter}, multi-view clustering with adaptive neighbours (MLAN) \cite{nie2017multi}, centroid-based co-regularized multi-view spectral clustering (CCo-MVSC)\cite{kumar2011co}, seven MF based methods (\emph{i.e.}, PMVC \cite{li2014partial}, IMG \cite{zhao2016incomplete}, IMC\_GRMF \cite{wen2018incompleteECCV}, MIC \cite{shao2015multiple}, OMVC \cite{shao2016online}, DAIMC \cite{hu2019doubly}, OPIMC \cite{hu2019one}), two kernel learning basd methods (\emph{i.e.}, CoKL \cite{shao2013clustering} and MKKM-IK-MKC \cite{liu2019multiple}), and two graph learning based methods (\emph{i.e.}, PIC \cite{wang2019spectral} and IMSC\_AGL \cite{wen2018incompleteTCYB}). MultiNMF is a popular non-negative MF multi-view clustering method. CCo-MVSC, AMGL, and MLAN are graph based multi-view clustering method, which require the full observation of views. For CCo-MVSC, the elements corresponding to the missing views in the graph are set as 0. For BSV, Concat, MultiNMF, AMGL, and and MLAN, the missing instances are filled in the average instance of the corresponding view. BSV performs kmeans on all views separately and then reports the highest clustering results. Concat stacks all views of a sample into a long vector and then performs kmeans on it to obtain the clustering results. For CoKL and MKKM-IK-MKC, we exploit the `Gaussian kernel' to construct the kernel matrices.
\subsection{Datasets}
In this subsection, we first list some public available multi-view datasets and their URLs for researchers. We divide the public datasets into two categories: 1) \textbf{Multiple-features}: In this category, some kinds of features are extracted from the same object such as document and image, as different views. Representative public multiple-features based multi-view datasets include: Handwritten digit images{\footnote{\url{https://archive.ics.uci.edu/ml/datasets/Multiple+Features}}}, Corel dataset for image retrieval{\footnote{\url{https://archive.ics.uci.edu/ml/datasets/Corel+Image+Features}}}, Caltech101 and NUSWIDE datasets for object recognition{\footnote{\url{https://drive.google.com/drive/folders/1O_3YmthAZGiq1ZPSdE74R7Nwos2PmnHH}}}, animal with attributes (AWA) dataset for animal classification{\footnote{\url{https://archive.ics.uci.edu/ml/datasets/Corel+Image+Features}}}, and some document datasets such as Cora, CiteSeer, WebKB, Newsgroup datasets{\footnote{\url{http://lig-membres.imag.fr/grimal/data.html}}} and BBCSport dataset{\footnote{\url{http://mlg.ucd.ie/datasets/bbc.html}}}.
2) \textbf{Multiple modalities}: Data in this category are collected from diverse domains or by different tensors. Some representative multiple modalities based multi-view datasets include: Columbia consumer video (CCV) dataset for consumer video/audio analysis{\footnote{\url{http://www.ee.columbia.edu/ln/dvmm/CCV/}}}, BUAA-visnir face dataset (BUAA){\footnote{\url{https://github.com/hdzhao/IMG/tree/master/data}}}, Berkeley Drosophila Genome Project (BDGP) for gene expression analysis{\footnote{\url{http://ranger.uta.edu/~heng/Drosophila/data/}}}, Reuters dataset for large-scale multilingual text analysis{\footnote{\url{https://archive.ics.uci.edu/ml/datasets/Corel+Image+Features}}}, and 3 Sources dataset{\footnote{\url{http://mlg.ucd.ie/datasets/3sources.html}}}. Among these datasets, BBCSport and 3 Sources datasets are naturally incomplete.

In our experiments, the following five representative public datasets are selected to compare different IMC methods, where their information are briefly summarized in Table I of the supplementary file:

(1) \textbf{\emph{BUAA}} \cite{huang2012buaa}: Face images in the BUAA dataset are collected by visual and near infrared cameras, which can be naturally regarded as two kinds of views of the same person. In our experiments, we select the subset composed of 90 visual and near infrared images from the first 10 classes as that in \cite{zhao2016incomplete} to evaluate the above represetnative IMC methods.

(2) \textbf{\emph{BDGP}} \cite{cai2012joint,tomancak2007global}: BDGP is designed for the research of gene expression. In our experiments, a multi-view BDGP dataset collected by Cai et al. \cite{cai2012joint} is chosen, where each sample is represented by the texture feature and three kinds of bag-of-words features extracted from the lateral, dorsal, and ventral images, respectively.

(3) \textbf{\emph{Caltech101}} \cite{fei2004learning}: Caltech101 is a popular object dataset which contains 9144 images from 102 categories including a background and 101 objects, such as airplanes, ant, bass, and beaver. In our experiments, four kinds of feature sets, \emph{i.e.}, Cenhist, Hog, Gist, and LBP, extracted by Li et al. are chosen as four views \cite{li2015large}.

(4) \textbf{\emph{BBCSport}} \cite{greene2006practical}:  BBCSport is a text dataset collected from the BBC Sport website corresponding to five kinds of sports news articles (\emph{i.e.}, athletics, cricket, football, rugby, and tennis) \cite{greene2006practical}. We select a subset{\footnote{\url{https://github.com/GPMVCDummy/GPMVC/tree/master/partialMV/PVC/recreateResults/data}}} which contains 116 samples represented by 4 views for evaluation \cite{rai2016partial}.

(5) \textbf{\emph{NUSWIDE}} \cite{chua2009nus}: NUSWIDE is a real-world web image dataset collected by the researchers from National University of Singapore. In our experiments, a multi-view subset containing 30,000 images and 31 classes is adopted, in which each image is represented by five kinds of low-level features, \emph{i.e.}, color histogram, color correlogram, edge direction histogram, wavelet texture, and block-wise color moments.
\subsection{Experimental setting}
\textbf{Evaluation Metrics}: Four well-known metrics, \emph{i.e.}, clustering accuracy (ACC), normalized mutual information (NMI), purity, and adjusted rand index (ARI), are chosen to assess the above methods \cite{liu2019multiple,amigo2009comparison,hubert1985comparing,zhang2015low}.

\textbf{Incomplete multi-view data construction}: For BUAA dataset, we randomly select $p\%$ samples as paired samples that have both two views, where $p$ is defined as \{10, 30, 50\}. For the remaining (1-$p\%$) samples, we remove the near infrared view for half of the samples, and remove the visual view for the other half. For the other four datasets, under the condition that all samples at least have one view, we randomly delete $p\%$ instances from every view to construct the incomplete data. Specifically, for each missing rate or paired rate $p\%$, we implement all compared methods on several groups of the randomly constructed incomplete data, then report the average and standard deviations of the clustering results.
\subsection{Experimental results and analysis}
\begin{table*}[t!]
\small
  \centering
  \caption{Experimental results \emph{w.r.t.} ACC (\%), NMI (\%), purity (\%) of different methods on the BUAA and BDGP datasets.}\resizebox{0.99\textwidth}{!}{
    \begin{tabular}{|c|l|ccc|ccc|ccc|}
    \hline
    \multicolumn{2}{|c|}{} & \multicolumn{3}{c|}{ACC}      & \multicolumn{3}{c|}{NMI}      & \multicolumn{3}{c|}{Purity} \\
    \hline
    \multicolumn{1}{|c|}{Data} & Method & 0.1   & 0.3   & 0.5   & 0.1   & 0.3   & 0.5   & 0.1   & 0.3    & 0.5   \\
    \hline
    \multirow{17}[2]{*}{\begin{turn}{-90}BUAA\end{turn}}
    &BSV	    &48.33$\pm$3.70	&56.96$\pm$4.93	&64.26$\pm$5.19	&43.10$\pm$3.77	&53.03$\pm$3.98	&61.78$\pm$3.85	&50.19$\pm$3.77 &58.66$\pm$3.46 &65.96$\pm$4.05\\
    &Concat	&45.62$\pm$2.54	&46.61$\pm$3.43	&47.46$\pm$3.29	&51.22$\pm$1.85	&51.95$\pm$2.78	&52.43$\pm$2.71	&47.99$\pm$2.30 &49.41$\pm$3.17 &49.68$\pm$3.16\\
    &MultiNMF &39.78$\pm$5.69	&44.44$\pm$6.09	&50.44$\pm$4.35	&40.96$\pm$4.86	&43.55$\pm$5.54	&49.03$\pm$5.95	&42.22$\pm$5.09 &46.67$\pm$5.50 &52.67$\pm$4.62\\
    &CCo-MVSC &56.74$\pm$4.91	&67.28$\pm$3.69	&72.60$\pm$3.46	&59.75$\pm$4.81	&69.36$\pm$2.94	&75.31$\pm$2.41	&59.29$\pm$4.74 &69.13$\pm$3.61 &74.90$\pm$3.19\\
    &AMGL &48.78$\pm$5.14	&55.11$\pm$4.92	&61.78$\pm$4.51	&46.93$\pm$4.22	&53.90$\pm$5.22	&59.24$\pm$2.87	&52.00$\pm$4.62	&58.11$\pm$4.48	&64.22$\pm$2.76  \\
    &MLAN &51.78$\pm$5.95	&51.44$\pm$5.78	&54.89$\pm$5.59	&52.49$\pm$5.09	&48.50$\pm$5.58	&48.75$\pm$6.17	&54.56$\pm$4.46	&53.67$\pm$5.11	&55.44$\pm$5.63\\
    &PMVC	&58.00$\pm$3.47	&65.02$\pm$5.99	&70.62$\pm$2.20	&60.40$\pm$3.25	&68.19$\pm$5.45	&72.86$\pm$2.43	&60.44$\pm$3.81 &67.16$\pm$5.46 &72.89$\pm$1.77\\
    &IMG	    &54.89$\pm$5.89	&67.39$\pm$5.79	&74.00$\pm$5.41	&53.46$\pm$5.55	&67.53$\pm$5.94	&76.26$\pm$3.53	&56.74$\pm$5.07 &68.65$\pm$3.46 &76.15$\pm$5.12\\
    &IMC\_GRMF	&66.89$\pm$7.43	&76.44$\pm$4.40	&82.44$\pm$2.88	&66.84$\pm$7.24	&76.91$\pm$4.37	&80.20$\pm$2.51	&69.11$\pm$5.47 &78.00$\pm$4.67 &82.44$\pm$2.88\\
    &MIC	        &56.58$\pm$4.48	&66.04$\pm$0.96	&71.02$\pm$8.61	&59.01$\pm$4.08	&68.39$\pm$0.44	&71.66$\pm$8.26	&59.11$\pm$4.53 &69.02$\pm$0.78 &73.02$\pm$8.02\\
    &OMVC	    &58.37$\pm$6.77	&62.67$\pm$5.10	&64.74$\pm$5.91	&59.41$\pm$5.40	&60.82$\pm$3.89	&63.89$\pm$4.96	&60.22$\pm$4.43 &63.41$\pm$5.02 &66.89$\pm$5.20\\
    &DAIMC	    &57.56$\pm$7.79	&65.33$\pm$7.30	&76.67$\pm$4.84	&57.65$\pm$7.30	&69.07$\pm$5.95	&80.62$\pm$2.99	&58.89$\pm$6.89 &69.11$\pm$6.40 &79.11$\pm$3.72\\
    &OPIMC	    &47.33$\pm$7.01	&52.67$\pm$4.13	&54.00$\pm$6.01 &49.42$\pm$6.12	&55.26$\pm$4.24	&55.16$\pm$7.62	&50.44$\pm$5.91 &55.33$\pm$4.87 &56.89$\pm$5.85\\
    &CoKL	    &21.33$\pm$0.93	&29.56$\pm$3.74	&38.89$\pm$3.96	&12.48$\pm$1.21	&22.48$\pm$2.54	&34.66$\pm$3.79	&23.78$\pm$0.99 &32.44$\pm$2.87 &41.73$\pm$4.49\\
    &MKKM-IK-MKC	&69.24$\pm$3.41	&\textbf{88.09$\pm$6.31}	&\textbf{91.11$\pm$3.86} &70.79$\pm$2.62	&\textbf{86.26$\pm$5.67}	&\textbf{88.98$\pm$3.77}	&70.89$\pm$2.70 &\textbf{88.09$\pm$6.31} &\textbf{91.11$\pm$3.86}\\
    &PIC	&64.44$\pm$6.82	&80.40$\pm$3.66	&90.44$\pm$2.68	&66.79$\pm$3.08	&80.99$\pm$1.78	&88.91$\pm$2.53	&67.16$\pm$5.93 &82.80$\pm$1.90 &90.44$\pm$2.67\\
    &IMSC\_AGL	&\textbf{74.89$\pm$6.41}	&81.11$\pm$5.21	&84.89$\pm$5.75	&\textbf{74.13$\pm$3.73}	&78.61$\pm$3.99	&82.80$\pm$4.96	&\textbf{75.89$\pm$4.90} &81.33$\pm$4.80 &84.89$\pm$5.75\\	
    \hline

    \multirow{13}[2]{*}{\begin{turn}{-90}BDGP\end{turn}}
    &BSV	&51.48$\pm$3.96	&41.44$\pm$3.55	&34.74$\pm$1.52	&35.74$\pm$4.01	&25.20$\pm$2.70	&16.39$\pm$1.36	&53.16$\pm$3.94	&43.75$\pm$2.63	&35.92$\pm$1.32\\
    &Concat	&57.66$\pm$4.79	&50.04$\pm$1.58	&40.41$\pm$3.52	&44.58$\pm$4.78	&31.81$\pm$1.45	&19.76$\pm$1.78	&59.68$\pm$4.71	&51.14$\pm$1.45	&41.81$\pm$2.79\\
    &MultiNMF &29.14$\pm$1.44	&30.39$\pm$1.33	&31.60$\pm$0.40	&7.97$\pm$0.90	&10.07$\pm$1.10	&8.54$\pm$0.65	&29.15$\pm$1.42	&30.39$\pm$1.33	&31.68$\pm$0.42\\
    &CCo-MVSC &61.50$\pm$5.64	&57.63$\pm$5.30	&52.03$\pm$3.40	&44.37$\pm$2.02	&35.78$\pm$2.58	&27.03$\pm$1.94	&64.84$\pm$3.83	&60.03$\pm$3.81	&53.93$\pm$2.36\\
    &AMGL &53.55$\pm$6.46	&44.99$\pm$6.67	&47.50$\pm$4.44	&44.63$\pm$7.98	&28.61$\pm$7.48	&\textbf{28.89$\pm$4.87}	&57.52$\pm$6.68	&46.62$\pm$6.64	&47.71$\pm$4.28\\
    &MLAN &32.24$\pm$4.64	&31.53$\pm$5.63	&27.37$\pm$1.62	&13.19$\pm$4.38	&12.46$\pm$5.28	&7.92$\pm$1.53	&32.84$\pm$4.14	&32.24$\pm$5.16	&27.90$\pm$1.49\\
    &MIC	&48.31$\pm$0.83	&40.88$\pm$1.18	&34.02$\pm$1.42	&28.52$\pm$0.49	&23.94$\pm$1.21	&11.05$\pm$0.89	&48.54$\pm$0.87	&43.57$\pm$1.05	&34.95$\pm$1.24\\
    &OMVC	&55.23$\pm$4.55	&46.22$\pm$3.15	&39.46$\pm$1.12	&28.78$\pm$1.59	&19.44$\pm$1.20	&13.51$\pm$1.21	&53.23$\pm$4.55	&46.34$\pm$2.89	&39.46$\pm$1.12\\
    &DAIMC	&77.34$\pm$2.58	&\textbf{69.30$\pm$6.42}	&52.45$\pm$8.57	&55.64$\pm$2.68	&\textbf{47.87$\pm$4.65}	&28.33$\pm$1.38	&\textbf{77.34$\pm$2.58}	&\textbf{69.34$\pm$6.41}	&52.88$\pm$7.73\\
    &OPIMC	&\textbf{79.38$\pm$7.69}	&63.73$\pm$7.29	&\textbf{55.17$\pm$9.24}&\textbf{61.77$\pm$7.41}	&41.47$\pm$3.88	&25.94$\pm$7.29	&79.38$\pm$7.69	&65.11$\pm$6.13	&\textbf{55.54$\pm$8.50}\\
    &MKKM-IK-MKC	&31.80$\pm$1.68	&29.12$\pm$0.29	&29.44$\pm$1.39	&7.21$\pm$1.10	&5.86$\pm$0.56	&6.65$\pm$1.19	&32.97$\pm$1.60	&30.15$\pm$0.64	&30.70$\pm$1.45\\
    &PIC	&32.92$\pm$9.14	&22.20$\pm$1.60	&23.52$\pm$0.56	&10.17$\pm$9.43	&1.97$\pm$1.58	&1.94$\pm$0.64	&33.16$\pm$9.28	&22.78$\pm$1.79	&23.95$\pm$0.81\\
    &IMSC\_AGL	&57.64$\pm$6.22	&52.08$\pm$9.20	&46.68$\pm$4.64	&34.77$\pm$3.96	&27.04$\pm$7.77	&21.50$\pm$3.50	&59.37$\pm$4.43	&53.29$\pm$8.28	&48.97$\pm$4.14\\
    \hline
    \end{tabular}}
  \label{tab:addlabel}%
\end{table*}
Experimental results in terms of ACC, NMI, and purity, of different methods on the BUAA and BDGP are listed in Tables I. For Caltech101, BBCSport, and NUSWIDE datasets, the experimental results of different methods are listed in Table II in the supplementary material owing to the space limitation. The results in terms of ARI are shown in Fig.1 in the supplementary material. It should be noted that since four graph learning based methods and a kernel learning based method, \emph{e.g.}, AMGL, MLAN, MKKM-IK-MKC, PIC, and IMSC\_AGL, require a very large memory and report `out of memory error' on our computer with 64GB RAM and Win 10 system, we do not report the experimental results of these methods on the large-scale NUSWIDE dataset in Table II and Fig.1 in the supplementary material. From the experimental results, we have the following observations:

(1) In most cases, all of the multi-view learning based IMC methods perform better than BSV and Concat on the first four datasets. In addition, the four complete multi-view clustering methods, \emph{i.e.}, MultiNMF, CCo-MVSC, AMGL, and MLAN, perform worse than the IMC methods, such as DAIMC, IMC\_GRMF, PIC, and IMSC\_AGL. These two phenomena demonstrate that: 1) all of the multi-view learning based IMC methods have the potential to capture more information from the incomplete multi-view data than BSV and Concat. 2) Simply setting the missing instances or graphs as 0 or average instance is not a good choice to address the incomplete clustering problem. 3) Sufficiently exploring the aligned information among the available views is an effective approach to address the incomplete learning problem for the difficult IMC tasks.

(2) On the BUAA, BBCSport, and Caltech101 datasets, PIC and IMSC\_AGL generally perform much better than the MF based IMC methods in terms of ACC, NMI, and purity. However, on the BDGP dataset, these two graph learning based methods and the kernel learning based method, \emph{e.g.}, MKKM-IK-MKC, perform worse than the other methods. Moreover, we can observe that PIC, IMSC\_AGR, and MKKM-IK-MKC are not suitable to the clustering task of the NUSWIDE dataset on a computer with 64 GB RAM memory. By analyzing the original features of data, we observe that BDGP is a very different dataset from the other four datasets since some original instances in this dataset are naturally unavailable and set as a zero vector. On this special dataset with some missing views, it is difficult to obtain the high quality kernels or graphs by `Gaussian kernel' or the distance based graph construction scheme. This is the major reason to the bad performance obtained by the kernel and graph based IMC methods. According to these phenomena, we can infer that: 1) in most cases, the graph learning based method can obtain a more discriminative representation than the MF based IMC methods for clustering. 2) Capturing the correct geometric structure information of data is very important for the unsupervised clustering tasks. 3) Compared with the MF based methods, the graph and kernel based methods need to compute several $n \ast n$ graph/kernel matrices for the data with $n$ samples, which require a large storage memory.

Moreover, from these observations, we can obtain that no IMC methods can maintain consistently good performance on all kinds of datasets. Therefore, it is significant to choose a suitable algorithm for different datasets.
\section{Conclusion}
The incomplete learning problem is challenging in multi-view clustering and its research is significant for practical applications. This paper reviewed almost all of the representative IMC methods and divided them into four categories, \emph{i.e.}, MF based IMC, kernel learning based IMC, graph learning based IMC, and deep learning based IMC. We not only briefly introduced some representative IMC methods, but also discussed their connections, differences, advantages and disadvantages in depth. For the MF based IMC methods, some unified frameworks were provided to integrate them.

Although many IMC methods have been proposed in the past decades, several challenging problems are still not solved well. For example:

\textbf{\emph{1) Large-scale problem}}. In practical clustering tasks, such as recommendation system and financial data analysis, the sample number of the collected data is often very large. However, most IMC methods suffer from a high computational complexity and memory cost, especially for the kernel based and graph based methods. As a result, the existing IMC methods are not applicable to large-scale datasets. Although some methods, such as OMVC and OPIMC, provide the chunk by chunk approach to reduce the memory requirement, the performance of these methods is not guaranteed. Therefore, it is significant to design some efficient methods that simultaneously consider the efficiency and performance for large-scale IMC tasks.

\textbf{\emph{2) Information imbalance problem}}. Owing to the random absence of views, different views often have different numbers of instances. In addition, different views not only commonly have huge differences in feature dimension and feature amplitude, but also describe the object at different levels. These indicate that different views carry different degrees of information for clustering. However, this factor has been consistently ignored by the existing IMC methods, which is not conducive to clustering. Therefore, designing a more robust clustering model by taking into account the information imbalance property is a new research direction for IMC.

\textbf{\emph{3) Mixed data types}} \cite{chao2017survey}. As far as we know, the existing IMC methods are all proposed for the multi-view data with vector based numerical features, but fails with data from other feature types, such as image, symbolic, and ordinal. Although one can preprocess the data to obtain the vector based features, some underlying information hidden in the data will be lost. For example, extracting the vector based features from images may miss some structure information. So it is worth to explore more applicable and flexible IMC models for mixed data.

\textbf{\emph{4) Complex data with noises}}. From the summarization to the existing IMC methods, it is easy to observe that existing methods all ignore the influence of noises since their models commonly treat all samples (including noisy samples and clean samples) equally important. As a result, the existing methods are not robust to noise. Generally, it is impossible to collect clean data without noise in practical applications. Thus, it is important to design more robust models that can reduce the negative influence of noises. In the single-view learning fields, self-paced learning has been proved an effective approach to identify the noisy samples and improve the robustness. Therefore, integrating the self-paced learning to design more robust models is a worth research direction.

\textbf{\emph{5) Missing view recovery}}. As far as we know, there are few works on missing view recovery. Moreover, the existing works suffer from some issues, where UEAF has a relatively high computational cost and the other two methods need sufficient samples whose all views are fully observed for model training. Moreover, the performances of missing-view recovery of these methods are neither guaranteed by experiments nor theory. In fact, missing view recovery not only can naturally solve the incomplete learning problem, but also is beneficial to improve the clustering performance. Besides this, missing view recovery has a lot of potential application values, especially in the criminal investigation. Therefore, it is of great significance to study the missing view recovery and design more robust unified missing-view recovery and IMC frameworks.

In addition to the above issues, some other issues existing in the multi-view clustering also appear in IMC, such as local minima and multiple solutions pointed out in \cite{chao2017survey}. Besides this, clustering on the partially view-aligned data \cite{huang2020partially,zhang2015constrained} is another challenging problem and the researches on both view-missing and view-partially-aligned cases are even rarer. In summary, the progress of IMC is still in the theoretical research stage. In the future, researchers need to design more efficient, high performance, and robust IMC methods for practical applications.

\bibliographystyle{IEEEtran}
\bibliography{IMC_survey}
\end{document}